\documentclass[letterpaper,11pt]{article}
\usepackage[utf8]{inputenc}
\usepackage[T1]{fontenc}
\usepackage[singlespacing]{setspace}
\usepackage[margin=1in]{geometry}
\usepackage{microtype}
\usepackage{needspace}

\usepackage{amsmath,amssymb,amsthm,mathtools}
\usepackage{thmtools}
\usepackage{thm-restate}
\usepackage{bm,bbm,nicefrac,pifont}
\usepackage{graphicx,xspace,xcolor,soul}
\sethlcolor{gray}

\usepackage[ruled,vlined]{algorithm2e}

\usepackage[numbers,sort]{natbib}
\usepackage{prettyref}
\usepackage{hyperref}
\hypersetup{
    colorlinks=true,
    linkcolor=blue,
    citecolor=blue,
    urlcolor=blue,
    pdftitle={Near-Optimal sqrt(K)-Regret for Online Learning in Uninformed Markov Games},
    pdfauthor={Junyan Liu, Haipeng Luo, Lillian J. Ratliff, Zihan Zhang}
}

\providecommand{\Aset}{\mathcal A}
\providecommand{\Bset}{\mathcal B}
\providecommand{\Sset}{\mathcal S}

\providecommand{\Prob}{\mathbb P}
\providecommand{\Indi}{\mathbf{1}}

\newcommand{\inner}[1]{ \left\langle {#1} \right\rangle }

\newcommand{\bi}{\begin{itemize}}
\newcommand{\ei}{\end{itemize}}

\newtheoremstyle{theorem}
    {3pt}
    {3pt}
    {\itshape}
    {}
    {\bfseries}
    {.}
    {.5em}
    {}
\theoremstyle{theorem} 
\newtheorem{theorem}{Theorem}[section]
\newtheorem{lemma}[theorem]{Lemma}

\DeclareMathOperator*{\argmin}{\arg\!\min}

\newcommand{\calJ}{\mathcal{J}}

\newcommand{\calV}{{\mathcal{V}}}

\newcommand{\field}[1]{\mathbb{#1}}

\newcommand{\E}{\field{E}}

\renewcommand{\P}{\field{P}}

\newcommand{\KL}{{\text{\rm KL}}}

\newcommand{\order}{\ensuremath{O}}
\newcommand{\otil}{\ensuremath{\widetilde{O}}}

\newcommand{\TV}{\texttt{TV}}

\newcommand{\ENR}{\textsc{Enr}}
\newcommand{\NR}{\textsc{Nr}}

\newcommand{\rbr}[1]{\left(#1\right)}

\newcommand{\sbr}[1]{\left[#1\right]}

\newcommand\numberthis{\addtocounter{equation}{1}\tag{\theequation}}

\DeclareFontFamily{OMX}{MnSymbolE}{}
\DeclareFontShape{OMX}{MnSymbolE}{m}{n}{
    <-6>  MnSymbolE5
   <6-7>  MnSymbolE6
   <7-8>  MnSymbolE7
   <8-9>  MnSymbolE8
   <9-10> MnSymbolE9
  <10-12> MnSymbolE10
  <12->   MnSymbolE12}{}
\DeclareSymbolFont{mnlargesymbols}{OMX}{MnSymbolE}{m}{n}
\SetSymbolFont{mnlargesymbols}{bold}{OMX}{MnSymbolE}{b}{n}
\DeclareMathDelimiter{\llangle}{\mathopen}{mnlargesymbols}{'164}{mnlargesymbols}{'164}
\DeclareMathDelimiter{\rrangle}{\mathclose}{mnlargesymbols}{'171}{mnlargesymbols}{'171}

\newcommand{\pref}[1]{\prettyref{#1}}

\newcommand{\savehyperref}[2]{\texorpdfstring{\hyperref[#1]{#2}}{#2}}
\newrefformat{eq}{\savehyperref{#1}{Eq.~\eqref{#1}}}
\newrefformat{eqn}{\savehyperref{#1}{Equation~\eqref{#1}}}
\newrefformat{lem}{\savehyperref{#1}{Lemma~\ref*{#1}}}
\newrefformat{def}{\savehyperref{#1}{Definition~\ref*{#1}}}
\newrefformat{line}{\savehyperref{#1}{line~\ref*{#1}}}
\newrefformat{thm}{\savehyperref{#1}{Theorem~\ref*{#1}}}
\newrefformat{corr}{\savehyperref{#1}{Corollary~\ref*{#1}}}
\newrefformat{cor}{\savehyperref{#1}{Corollary~\ref*{#1}}}
\newrefformat{col}{\savehyperref{#1}{Corollary~\ref*{#1}}}
\newrefformat{sec}{\savehyperref{#1}{Section~\ref*{#1}}}
\newrefformat{subsec}{\savehyperref{#1}{Section~\ref*{#1}}}
\newrefformat{app}{\savehyperref{#1}{Appendix~\ref*{#1}}}
\newrefformat{assum}{\savehyperref{#1}{Assumption~\ref*{#1}}}
\newrefformat{ex}{\savehyperref{#1}{Example~\ref*{#1}}}
\newrefformat{fig}{\savehyperref{#1}{Figure~\ref*{#1}}}
\newrefformat{tab}{\savehyperref{#1}{Table~\ref*{#1}}}
\newrefformat{alg}{\savehyperref{#1}{Algorithm~\ref*{#1}}}
\newrefformat{rem}{\savehyperref{#1}{Remark~\ref*{#1}}}
\newrefformat{conj}{\savehyperref{#1}{Conjecture~\ref*{#1}}}
\newrefformat{prop}{\savehyperref{#1}{Proposition~\ref*{#1}}}
\newrefformat{proto}{\savehyperref{#1}{Protocol~\ref*{#1}}}
\newrefformat{prob}{\savehyperref{#1}{Problem~\ref*{#1}}}
\newrefformat{claim}{\savehyperref{#1}{Claim~\ref*{#1}}}
\newrefformat{que}{\savehyperref{#1}{Question~\ref*{#1}}}
\newrefformat{obs}{\savehyperref{#1}{Observation~\ref*{#1}}}

\providecommand{\Vset}{\mathcal V}

\providecommand{\term}{S_{H+1}}

\title{Near-Optimal $\sqrt{K}$-Regret for Online Learning \\ in Uninformed Markov Games\thanks{Authors are listed in alphabetical order.}}

\author{%
Junyan Liu\thanks{University of Washington. Email: \texttt{junyanl1@cs.washington.edu}.}
\and
Haipeng Luo\thanks{University of Southern California. Email: \texttt{haipengl@usc.edu}.}
\and
Lillian J. Ratliff\thanks{University of Washington. Email: \texttt{ratliffl@uw.edu}.
}
\and
Zihan Zhang\thanks{Hong Kong University of Science and Technology. Email: \texttt{zihanz@ust.hk}.
}
}

\date{}

\begin{document}
\pagenumbering{roman}
\maketitle
\thispagestyle{empty}

\begin{abstract}
We study online learning in two-player uninformed Markov games, where the opponent may follow an arbitrary, unknown policy in each episode and the learner does not observe the opponent's actions. \cite{tian2021online} showed that achieving sublinear external regret requires an exponential dependence on the episode length $H$. Motivated by this limitation, they studied the weaker notion of Nash-value regret and obtained an $\widetilde{O}(K^{2/3})$ regret bound, where $K$ is the number of episodes. More recently, \cite{liu2026online} introduced a stronger notion called empirical Nash-value regret and achieved an $\widetilde{O}(\sqrt{K})$ bound when the opponent's non-stationarity is moderate, including the case where the opponent follows a fixed policy. However, whether the $\sqrt{K}$ rate is achievable in the worst case remained open.

In this paper, we resolve this open question by developing an algorithm that achieves $\widetilde{O}(H^{3/2}S\sqrt{AK})$ empirical Nash-value regret against \textit{arbitrary} opponent policies, where $S$ is the number of states and $A$ is the number of learner's actions.
Specifically, our algorithm combines global and local planning by maintaining a collection of candidate predictions of future losses, each associated with a policy. Global planning learns which candidates to favor across episodes, while local planning combines their policies differently at each state to make decisions. A key ingredient is to coordinate these two components to control the discrepancy between local loss estimates and the total loss of the executed policy, using only the learner's observed trajectories.

\end{abstract}

\clearpage
\tableofcontents
\thispagestyle{empty}

\clearpage
\pagenumbering{arabic}

\section{Introduction}
\label{sec:intro}

Sequential decision making often takes place in environments where outcomes
depend on the behavior of other agents. Examples include interactive
autonomous driving
\citep{kiran2022autonomous,yu2020distributed},
wireless resource allocation
\citep{luong2019communications,naderializadeh2021resource},
online advertising and repeated auctions
\citep{weed2016auctions,balseiro2019auctions},
and security resource allocation
\citep{jain2010security,an2012security},
where the effect of one agent's decision depends on how other agents respond.
Markov games (MGs) provide a basic model for such interactions by combining
strategic decisions with state-dependent dynamics
\citep{shapley1953stochastic}.

Many existing works on MGs study the self-play setting, where
both players are controlled by the same learning procedure and their actions are typically observable
\citep{bai2020provable,bai2020near,liu2021sharp}.
In many applications, however, the learner interacts with independently controlled agents whose policies may be arbitrary and may change over time, while their actions are only partially observed or entirely unobserved.
For example, in distributed wireless resource allocation, multiple
transmitters make simultaneous decisions without knowing the concurrent decisions of other transmitters
\citep{naderializadeh2021resource}, while security applications may involve
strategic interactions under limited surveillance
\citep{an2012security}. These information constraints motivate the
\textit{uninformed MGs} setting, where the learner interacts with an
arbitrary opponent without observing the opponent's actions
\citep{tian2021online,liu2026online}.

Unlike informed MGs, where sublinear external-regret guarantees have been
well established, \cite{tian2021online} show that sublinear external regret
in uninformed MGs cannot be achieved without an exponential dependence on the
episode length $H$. Motivated by this negative result, they consider the
weaker notion of Nash-value regret ($\NR$), which compares the learner's
performance with the minimax value of the underlying game, and establish an
$\widetilde O(K^{2/3})$ bound, where $K$ is the number of episodes.
More recently, \cite{liu2026online} introduce empirical Nash-value regret
($\ENR$), a stronger benchmark than $\NR$ that recovers standard external
regret when the opponent follows a fixed policy. They develop an algorithm that adapts to the opponent's non-stationarity, achieving an $\widetilde O(\sqrt K)$ rate for a fixed opponent while retaining an $\widetilde O(K^{2/3})$ rate in the worst case.
Despite such progress, a natural question remains:
\begin{center}
\emph{Can one achieve $\ENR_K=\widetilde O(\sqrt K)$
guarantee in the worst case for uninformed MGs?}
\end{center}

\textbf{Our contributions.} In this work, we answer this question affirmatively. We develop a new
algorithm for uninformed MGs and prove that, against an arbitrary adaptive
opponent, it achieves
\[
\ENR_K=\widetilde O\!\left(H^{3/2}S\sqrt{AK}\right)
\]
with high probability, where $S$ and $A$ denote the numbers of states and actions, respectively. To the best of our knowledge, this is the first
$\widetilde O(\sqrt K)$ worst-case guarantee for (empirical) Nash-value regret
in uninformed MGs. It improves the previous worst-case dependence on the
number of episodes from $K^{2/3}$ to $\sqrt K$, matching the optimal dependence on $K$ and $A$ implied by
the $\Omega(H\sqrt{SAK})$ lower bound for episodic
MDPs~\cite{domingues2021episodic}.\footnote{
Since $\ENR_K\ge\NR_K$ and episodic MDPs are special cases of uninformed
MGs with a single opponent action, the MDP lower bound also applies
to $\ENR_K$.
} 
While this establishes the statistical achievability of the optimal $\sqrt{K}$ rate, our algorithm is not computationally
efficient, and achieving the same guarantee efficiently
remains an interesting open problem.

\subsection{Preliminaries}
\label{sec:setting}

We study an episodic two-player Markov game with $K\ge1$ episodes of
$H\ge1$ steps/stages each. At each step $h\in\{1,\ldots,H\}$, the game has a
finite state space $\Sset_h$ and finite action spaces $\Aset_h$ and
$\Bset_h$ for the learner and the opponent, respectively.
For simplicity, states at different steps belong to disjoint
sets, so $\Sset=\bigcup_{h=1}^H\Sset_h$ contains $S=|\Sset|$ nonterminal
states, with $\Sset_{H+1}=\{\term\}$ containing the terminal state.
The overall action spaces are $\Aset=\bigcup_{h=1}^H\Aset_h$ and
$\Bset=\bigcup_{h=1}^H\Bset_h$, with $A=|\Aset|$.
The fixed, unknown transition and loss functions are
$P_h\colon\Sset_h\times\Aset_h\times\Bset_h\to\Delta(\Sset_{h+1})$ and
$\ell_h\colon\Sset_h\times\Aset_h\times\Bset_h\to[0,1]$, respectively,
where $\Delta(X)$ denotes the set of probability distributions over $X$.
The learner knows $H$, $K$, the state spaces, and its own action spaces, but not
the opponent's action spaces.

At the beginning of each episode $k$, the learner and opponent commit to
Markov policies $\mu^k=\{\mu_h^k\}_{h=1}^H$ and
$\nu^k=\{\nu_h^k\}_{h=1}^H$, respectively.
These policies remain fixed throughout the episode, with
$\mu_h^k(\cdot\mid s)\in\Delta(\Aset_h)$ and
$\nu_h^k(\cdot\mid s)\in\Delta(\Bset_h)$ for every $s\in\Sset_h$.
The episode starts from an adversarially chosen initial state
$s_1^k\in\Sset_1$. At step $h$, the players observe $s_h^k$ and
simultaneously draw $a_h^k\sim\mu_h^k(\cdot\mid s_h^k)$ and
$b_h^k\sim\nu_h^k(\cdot\mid s_h^k)$, independently conditional on the
history and current state.
The learner then incurs loss $\ell_h^k=\ell_h(s_h^k,a_h^k,b_h^k)$,
and the environment transitions to
$s_{h+1}^k\sim P_h(\cdot\mid s_h^k,a_h^k,b_h^k)$.
The learner observes only its own action, the loss, and the next state,
while the opponent's action and policy remain unobserved.
The opponent may adapt its policy across episodes using the interaction
history, including the learner's past actions, and may choose $\nu^k$ after observing the learner's policy $\mu^k$.
It may know the game and the learner's algorithm, but cannot anticipate
future random draws by the learner or the environment.

For a policy pair $(\mu,\nu)$, step $h$, and state $s\in\Sset_h$, the state
value function is the expected cumulative loss from $s$ through step $H$,
\[
 V_h^{\mu,\nu}(s)
 =\E_{\mu,\nu}\left[
   \sum_{h'=h}^H\ell_{h'}(s_{h'},a_{h'},b_{h'})
   \,\middle|\,s_h=s
 \right],
\]
where the expectation is over the trajectory generated by $(\mu,\nu)$,
and $V_{H+1}^{\mu,\nu}(\term)=0$.
For any function $V$ on $\Sset_{h+1}$, we use the shorthand
$P_hV(s,a,b)=\sum_{s'\in\Sset_{h+1}}P_h(s'\mid s,a,b)V(s')$.
The value function then satisfies the Bellman equation
$V_h^{\mu,\nu}(s)=
\E_{\substack{a\sim\mu_h(\cdot\mid s),b\sim\nu_h(\cdot\mid s)}}
\bigl[\ell_h(s,a,b)+P_hV_{h+1}^{\mu,\nu}(s,a,b)\bigr]$.

Following \cite{liu2026online}, we measure performance by empirical
Nash-value regret ($\ENR$).
For the realized opponent policies $\nu^1,\ldots,\nu^K$, the empirical
Nash value is defined recursively at each step $h$ and state $s\in\Sset_h$ by
\begin{equation}
\begin{aligned}
 V_h^{\mathrm{emp}}(s)
 &={}\min_{x\in\Delta(\Aset_h)}
 \max_{\nu\in\{\nu_h^1(\cdot\mid s),\ldots,\nu_h^K(\cdot\mid s)\}}
 \sum_{a\in\Aset_h}\sum_{b\in\Bset_h}
 x(a)\nu(b)
 \bigl[\ell_h(s,a,b)+P_hV_{h+1}^{\mathrm{emp}}(s,a,b)\bigr],
\end{aligned}
 \label{eq:empirical-value}
\end{equation}
with $V_{H+1}^{\mathrm{emp}}(\term)=0$.
Empirical Nash-value regret ($\ENR$) is defined as
\begin{equation}
 \ENR_K=\sum_{k=1}^K
 \left[V_1^{\mu^k,\nu^k}(s_1^k)-V_1^{\mathrm{emp}}(s_1^k)\right].
 \label{eq:empirical-regret}
\end{equation}

Allowing $\nu$ to range over all of $\Delta(\Bset_h)$ in
\pref{eq:empirical-value} defines the standard Nash value $V_h^\star(s)$.
Let $(\mu^\star,\nu^\star)$ denote a pair of Nash Markov policies
for the learner and the opponent, respectively, such that
$(\mu_h^\star(\cdot\mid s),\nu_h^\star(\cdot\mid s))$ forms a saddle point
for the one-step loss $\ell_h(s,a,b)+P_hV_{h+1}^\star(s,a,b)$
at every $(h,s)$. Consequently,
$V_h^\star(s)=V_h^{\mu^\star,\nu^\star}(s)$.
Since the empirical benchmark restricts the opponent's choices,
$V_h^{\mathrm{emp}}(s)\le V_h^\star(s)$, so $\ENR_K$ is
a stronger performance measure than Nash-value regret
$\NR_K=\sum_{k=1}^K
\big[V_1^{\mu^k,\nu^k}(s_1^k)-V_1^\star(s_1^k)\big]$.
Moreover, when the opponent follows a fixed policy
$\nu^1=\cdots=\nu^K$, the empirical value reduces to
$V_h^{\mathrm{emp}}(s)=\min_\mu V_h^{\mu,\nu}(s)$.
Consequently, $\ENR_K$ recovers standard external regret.

\textbf{Additional notations.}
For a positive integer $n$, let $[n]=\{1,\ldots,n\}$.
In sums and products, unspecified index ranges are $h\in[H]$,
$s\in\Sset_h$, $a\in\Aset_h$, and $b\in\Bset_h$.
We write $[x]_+=\max\{x,0\}$ for the positive part of $x$.
For probability distributions $p,q$ on a finite set,
$\KL(p\Vert q)=\sum_i p_i\log(p_i/q_i)$ denotes the
Kullback--Leibler divergence.
Their total variation distance is
$\TV(p,q)=\frac12\sum_i|p_i-q_i|$.
For a differentiable convex function $\mathcal H$,
its Bregman divergence is defined as
$D_{\mathcal H}(u'\Vert u)
=\mathcal H(u')-\mathcal H(u)
-\langle\nabla\mathcal H(u),u'-u\rangle$,
for any $u,u'$ in its domain.
For each episode $k$, let $\mathcal F^k$ contain the
interaction history, the initial state $s_1^k$, and
both committed policies $\mu^k,\nu^k$, before the
trajectory is sampled.
We write $\E^k[\cdot]=\E[\cdot\mid\mathcal F^k]$ and
$\Prob^k(E)=\Prob(E\mid\mathcal F^k)$ for any event $E$.
For each step $h$, let $\mathcal F_h^k$ additionally
contain the observations up to the current state
$s_h^k$, before the step-$h$ actions are sampled,
and write $\E_h^k[\cdot]=\E[\cdot\mid\mathcal F_h^k]$.

\subsection{Limitations of Prior Approaches}
\label{subsec:prior-limitations}

Both~\cite{tian2021online} and~\cite{liu2026online} adopt a stepwise V-learning framework, where each state--stage pair maintains a local bandit subroutine that learns from immediate losses and downstream value estimates. The two approaches differ mainly in how they incorporate historical feedback: \cite{tian2021online} gradually downweight older observations, whereas~\cite{liu2026online} employ epoch-based updates and periodically restart local bandits.
However, such stepwise approaches face an intrinsic difficulty in achieving a worst-case regret rate better than $\widetilde{O}(K^{2/3})$. We illustrate this difficulty through the epoch-based V-learning algorithm of~\citep{liu2026online}, for which we establish the following lower bound. See \pref{app:epoch-v-learning-lb} for the proof.

\begin{restatable}{proposition}{LBVlearning}
\label{prop:epoch-v-learning-lb}
For any $K\ge 1$ and fixed $\eta\in(0,1/2]$, there exist a two-player
uninformed Markov game with $H=2$ and constant-sized state and action spaces,
and an oblivious opponent that changes its policy at most once, such that
the fixed-$\eta$ epoch V-learning algorithm of~\citep{liu2026online} satisfies $
    \mathbb{E}[\NR_K] = \Omega(K^{2/3}).$
\end{restatable}

The key intuition in this lower bound is that historical value estimates can lag behind changes in the opponent's behavior. When the loss at a second-step state increases, earlier low-loss observations keep its estimate below the current loss, even if it is updated after every visit. The first-step learner therefore receives feedback that understates the loss of its executed policy, and this discrepancy can accumulate over time.

For example, consider a
two-step game in which the learner visits a second-step state $s$
in every round and has only one action there. 
Suppose the opponent induces zero loss for the first $K-K^{2/3}$
rounds and unit loss for the remaining $K^{2/3}$ rounds. After $j$
rounds in the latter phase, the running average is 
$j/(K-K^{2/3}+j)\leq K^{-1/3}$.  The cumulative estimated loss passed to the first step is thus
at most $K^{1/3}$, whereas the cumulative realized loss is
$K^{2/3}$, yielding a discrepancy of order $K^{2/3}$.

Giving more weight to recent observations can reduce this lag, but it also reduces the effective sample size and makes the estimate more sensitive to noise. To capture the cost of forgetting, the proof uses a complementary stationary noisy instance, in which frequent restarts
erase information needed to distinguish actions with small loss gaps. For the fixed-parameter epoch scheme, these two instances expose competing costs: informally, an effective epoch length of $W$ leads to a cost of order $W$ from slow adaptation and $K/\sqrt{W}$ from repeatedly learning the local decisions. Balancing these costs gives $W\asymp K^{2/3}$ and a worst-case cost of order $K^{2/3}$. 

 This lower bound highlights a fundamental difficulty arising from
the different scopes of local and global planning. Local planning
estimates the remaining loss at each $(h,s)$, whereas
global planning evaluates the loss of the policy executed over
the entire episode. Since local estimates are learned and updated
independently, they need not be consistent with the global
roll-out loss, especially when the opponent changes its behavior.
The proposition illustrates how this discrepancy can lead to
substantial regret, motivating us to explicitly control the
gap between local feedback and global roll-out loss in
\pref{subsec:technical-ingredients}.

\subsection{Overview of Our Algorithm}
\label{subsec:algorithm-overview}

Our algorithm maintains a finite class $\mathcal V\subset \mathbb{R}^{S}$ of candidate loss tables with $N=|\mathcal V| \leq (HK+1)^S$, where each candidate $v = \bigl\{v_h(s)\bigr\}_{h\in[H],\,s\in\mathcal S_h}
\in \mathbb R^S$ specifies a prediction $v_h(s)$ of the normalized empirical Nash value  $V_h^{\mathrm{emp}}(s)/H$.

In particular, we set $v_{H+1}(\term)=0$ and require
$v_h(s)\in\bigl[0,\frac{H-h+1}{H}\bigr]$ for $2\le h\le H$. We fix $v_1(s)=0$ because a separate initial-value prediction is unnecessary: a low predicted initial loss is meaningful only when
supported by downstream predictions, which provide the basis for evaluating initial performance. Predictions at later steps remain necessary because they inform decisions at the preceding step.

At the beginning of episode $k$, the learner maintains a distribution
$w^k\in\Delta(\mathcal V)$ representing its global belief over the candidate
loss functions. For every candidate $v$ and $(h,s)$ pair,
it also learns a separate action distribution $\pi_h^k(\cdot\mid s,v)$.
This local policy reflects the learner's preferences among actions at
$(h,s)$ under the hypothesis that $v$ is the optimal cumulative loss. Together, $w^k$ and $\pi^k$ retain the information accumulated
across episodes about the candidates and the actions associated with them.

To construct its policy in episode $k$, the learner uses the global belief
$w^k$ and the current local policies $\pi^k$ to search for locally calibrated
mixing weights $q_h^k(\cdot\mid s)\in\Delta(\mathcal V)$ at each location
$(h,s)$. It then executes the mixture policy
\[
    \mu_h^k(a\mid s)
    =
    \sum_{v\in\mathcal V}
    q_h^k(v\mid s)\pi_h^k(a\mid s,v).
\]

This design reflects the distinct scopes of local action selection and global candidate assessment, together with the need for consistency across steps. By the Markov property, the mixed policy at $(h,s)$ is evaluated through the remaining loss from that state, whereas candidates are assessed globally through the total loss from the initial state. This difference motivates allowing the candidate weights to vary across $(h,s)$ pairs. The question is then how to choose these state-dependent weights. A key consideration is to align the local feedback used for learning with the total loss of the executed policy. This requires accounting for the coupling of loss predictions across steps: 
a value assigned to a downstream state enters the loss prediction at an upstream state, and trajectory feedback is used to assess whether the associated policies collectively support these predictions. The optimization over $q^k$ therefore calibrates local policy execution using both the global candidate weights and the consistency of predictions across steps.

After each episode, trajectory feedback jointly updates $w^k$ and $\pi^k$ through mirror descent. We design the mixing weights and this joint update together so that progress in learning the candidate weights and their associated policies remains aligned with the loss of the executed mixture.

\subsection{Key Techniques and Analysis behind Algorithm}
\label{subsec:technical-ingredients}

To illustrate the key ideas, we assume full-information access to the one-step feedback and sketch a simplified proof of the expected regret
bound $\E[\NR_K]=\otil\bigl(HS^{3/2}\sqrt{K}\bigr)$.
Compared with our main high-probability bound
$\ENR_K=\widetilde O\!\left(H^{3/2}S\sqrt{AK}\right)$,
this bound saves a factor of $\sqrt{A}$ through the full-information assumption, but incurs an additional factor of $\sqrt{S/H}$ due to a
suboptimal choice of parameters.
For this proof sketch, we assume that $\calV$ contains a candidate $v^\star$ satisfying $v_h^\star(s)=V_h^\star(s)/H$ for $2\le h\le H$, with $v_1^\star(s)=v_{H+1}^\star(\term)=0$.

\paragraph{Update $\pi^k_h,w^k$ via joint online mirror descent.}

Learning the candidate weights $w^k$ and the candidate-specific action distributions $\pi^k$ involves two coupled tasks. The weight $w^k(v)$ represents the learner's global belief in candidate $v$,
while the associated policy $\pi_h^k(\cdot\mid s,v)$ captures the action
preferences learned at $(h,s)$ under the continuation-loss predictions
specified by $v$.
Analyzing their updates separately would therefore require controlling the interactions between these two learning tasks. To address this coupling, we introduce the joint variable $z_{h,s,v,a}^k=w^k(v)\pi_h^k(a\mid s,v)$ and the feasible set as
\begin{equation}
\label{eq:ti-joint-domain}
\mathcal Z=
\left\{
z = \{z_{h,s,v,a}\}_{h\in [H],s\in \mathcal{S}_h,v\in \mathcal{V},a\in \mathcal{A}_h}\ge 0:
\exists w\in\Delta(\mathcal V),\
\sum_{a\in\mathcal A_h}z_{h,s,v,a}=w(v),
\quad\forall h,s,v
\right\}.
\end{equation}

Each $z\in \mathcal{Z}$ naturally introduces a weight $w$ and a policy $\pi$ as 
\begin{align}
w(v) = \sum_{a} z_{h,s,v,a},\qquad \pi_h(a|s,v)= \frac{z_{h,s,v,a}}{w(v)}\quad\text{if }w(v)>0.\label{eq:induce}
\end{align}
When $w(v)=0$, $\pi_h(\cdot\mid s,v)$ is any distribution on $\Aset_h$.
For each $(h,s)$ pair, $z_{h,s,\cdot,\cdot}$ is a joint distribution over candidate--action pairs. All states share the same candidate marginal $w$, making $\mathcal Z$ a convex polytope.
We apply online mirror descent (OMD) to update $z^k$ using the negative-entropy regularizer
\begin{align}
\mathcal H(z)=\sum_{h,s,v,a}z_{h,s,v,a}\log z_{h,s,v,a} = S\sum_{v}w(v) \log(w(v)) + \sum_{h,s,v,a}z_{h,s,v,a}\log \frac{z_{h,s,v,a}}{w(v)} \label{eq:defent}
\end{align}
and its Bregman divergence $D_{\mathcal H}$, with zero-mass entropy terms interpreted as zero. For a learning rate $0<\lambda\le1$, the update starts from a uniform distribution and takes the form
\begin{align}
z^1_{h,s,v,a} & =\frac{1}{N|\Aset_h|}, \quad \forall h\in [H], s\in \mathcal{S}_h, v\in \mathcal{V}, a\in \mathcal{A}_h\label{eq:ti-mirror-step}
\\z^{k+1}
 & =\operatorname*{arg\,min}_{z\in\mathcal Z}
\left\{ \lambda
\langle z,g^k\rangle
+D_{\mathcal H }(z\Vert z^k)
\right\}.\nonumber
\end{align}
Here, \(g^k=\{g^k_{h,s,v,a}\}_{h,s,v,a}\) denotes the feedback received in episode \(k\), with action block $g_{h,s,v}^k=(g_{h,s,v,a}^k)_{a\in\Aset_h}$. We will later specify \(g^k\) so that the resulting OMD updates can be aligned with the roll-out value. The updated candidate weights $w^{k+1}$ and policy $\pi^{k+1}$ are recovered from $z^{k+1}$ following \pref{eq:induce}.

The joint OMD framework allows us to compare the learner with the optimal candidate--policy pair. The comparison point $z^\star_{h,s,v,a}=\Indi\{v=v^\star\}\mu_h^\star(a\mid s)$ assigns all candidate weight to $v^\star$ and follows its Nash policy. Let $L^k=\mathbb E^k\!\left[
\sum_{h=1}^H\frac{\ell_h^k}{H}
 \right]$ denote the executed policy's normalized expected episode loss. Adding and subtracting the learner's and comparator's joint feedback losses yields
\begin{align}
\frac{\NR_K}{H}
&=\sum_{k=1}^K
\left(L^k-\frac{V_1^\star(s_1^k)}{H}\right)
\nonumber\\
&=\underbrace{\sum_{k=1}^K
\langle z^k-z^\star,g^k\rangle}_{\text{OMD regret}}
+\underbrace{\sum_{k=1}^K
\left(L^k-\langle z^k,g^k\rangle\right)}_{\text{drift error}}
+\underbrace{\sum_{k=1}^K
\left(\langle z^\star,g^k\rangle
-\frac{V_1^\star(s_1^k)}{H}\right)}_{\text{optimality gap}}.
\label{eq:reg_decom}
\end{align}

The first term measures the learner's regret against the optimal candidate--policy pair, the drift error captures the discrepancy between this feedback and the loss of the executed mixture policy, and the optimality gap relates the comparator's feedback to the Nash value benchmark.

\subsubsection{Design of Feedback $g^k$.}

We choose the feedback $g^k$ as a scaled Bellman residual so that the optimality gap in \pref{eq:reg_decom} is non-positive in expectation. Specifically,
\begin{equation}
\label{eq:defgk}
g^k_{h,s,v,a}
=\frac{  q^k_h(v|s)\Indi\{s_h^k=s\} }{ w^k(v) }
\left( Y_{h,s,v,a}^k-v_h(s)\right),
\end{equation}
where $
Y_{h,s,v,a}^k
=\sum_{b\in\Bset_h}\nu_h^k(b\mid s)\left[    \ell_h(s,a,b)/H + P_h v_{h+1}(s,a,b)\right]$
denotes the expected sum of the normalized immediate loss and the future loss predicted by candidate $v$. For simplicity, we assume full-information access to
$Y_{h,s,v,a}^k$ in this proof sketch; in the full algorithm,
we replace it with an empirical estimator constructed from the observed
trajectory.

 At a high level, the one-step feedback
 $Y_{h,s,v,a}^k$ evaluates action $a$ under the hypothesis
that $v$ gives normalized Nash value function. It combines the immediate loss with the loss predicted by $v_{h+1}$. The residual $Y_{h,s,v,a}^k-v_h(s)$ then assesses the consistency of $v$ across successive steps. A large positive residual suggests that $v_h(s)$ underestimates
the immediate loss plus the predicted remaining loss, providing evidence for reducing the global credibility of $v$. 

As discussed in \pref{subsec:algorithm-overview}, we introduce locally calibrated
mixing weights $q_h^k(\cdot \mid s)$ to control the discrepancy between global
and local planning. Specifically, the ratio $q_h^k(v \mid s)/w^k(v)$ acts as a
change-of-measure factor, reweighting the OMD feedback to align with the
executed policy.
To see this, consider a visited pair $(h,s)$ and write
$f(v)=\sum_a\pi_h^k(a\mid s,v)
(Y_{h,s,v,a}^k-v_h(s))$
for the Bellman residual averaged over the action
distribution associated with candidate $v$.
Since $z_{h,s,v,a}^k=w^k(v)\pi_h^k(a\mid s,v)$,
OMD naturally weights these residuals according to
$w^k$, whereas the executed policy uses the local
candidate weights $q_h^k(\cdot\mid s)$.
Without the ratio, the OMD feedback would therefore
average $f(v)$ under $w^k$, rather than $q_h^k$.
Multiplying by $q_h^k(v\mid s)/w^k(v)$ converts
one weighted average into the other:
\begin{align*}
\mathbb E_{v\sim w^k}
\left[
\frac{q_h^k(v\mid s)}{w^k(v)}f(v)
\right]
=
\mathbb E_{v\sim q_h^k(\cdot\mid s)}
[f(v)].
\end{align*}
This is precisely the change-of-measure identity
from $w^k$ to $q_h^k(\cdot\mid s)$.
Thus, although OMD maintains the global candidate
weights $w^k$, its feedback evaluates the candidate
Bellman residuals using the same local weights
$q_h^k$ that determine the executed mixture policy. 

\subsubsection{Bounding Optimality Gap}

Since $\mu^\star$ is a Nash policy, for every $h\ge2$ and state $s$, we have
\begin{align*}
\sum_a\mu_h^\star(a\mid s)Y_{h,s,v^\star,a}^k
&=
\mathbb E_{\substack{
a\sim\mu_h^\star(\cdot\mid s),\,b\sim\nu_h^k(\cdot\mid s)\\
s'\sim P_h(\cdot\mid s,a,b)
}}
\left[
\frac{\ell_h(s,a,b)}{H}+v_{h+1}^\star(s')
\right]\\
&\le
\max_{b\in\mathcal B_h}
\mathbb E_{\substack{
a\sim\mu_h^\star(\cdot\mid s)\\
s'\sim P_h(\cdot\mid s,a,b)
}}
\left[
\frac{\ell_h(s,a,b)}{H}+v_{h+1}^\star(s')
\right] \\
&=v_h^\star(s).
\end{align*}

Thus, the expected Bellman residual of the Nash comparator is nonpositive at every noninitial state, regardless of the opponent's committed policy. This implies
\begin{align*}
\mathbb E^k\!\left[\langle z^\star,g^k\rangle\right]
&=
\mathbb E^k\!\left[
\sum_{h=1}^H
\frac{  q^k_h(v^\star|s_h^k) }{ w^k(v^\star) }
\left(
\sum_a\mu_h^\star(a\mid s_h^k)
Y_{h,s_h^k,v^\star,a}^k
-v_h^\star(s_h^k)
\right)
\right]\\ \nonumber
&\le
\mathbb{E}^k\left[\frac{  q^k_1(v^\star |s_1^k) }{ w^k(v^\star) }
\sum_a\mu_1^\star(a\mid s_1^k)
Y_{1,s_1^k,v^\star,a}^k \right]\\ \nonumber
&\le
\mathbb{E}^k\left[\frac{  q^k_1(v^\star |s_1^k) }{ w^k(v^\star) }\cdot 
\frac{V_1^\star(s_1^k)}{H}\right],
\end{align*}
where the first equality follows from  the definition of $z^\star$ and $g^k$, the first inequality uses the non-positive Bellman residuals for $h\ge2$ and $v_1^\star(s)=0$, the second follows from the Nash Bellman inequality at step~1.

By setting $q^k_1(v|s)=w^k(v)$ for all $s\in\mathcal S_1$ and $v\in \mathcal{V}$, we learn that the conditional expected optimality gap is non-positive: $\mathbb{E}^k \left[\langle z^\star,g^k\rangle-V_1^\star(s_1^k)/H\right]   \leq 0$.

\subsubsection{Bounding OMD Regret.}
For a suitable choice of step size, the standard OMD analysis yields
\begin{equation*}
\langle z^k-z^\star,g^k\rangle
\le
\frac{D_{\mathcal H}(z^\star\Vert z^k)
-D_{\mathcal H}(z^\star\Vert z^{k+1})}{\lambda}
+\lambda\calJ^k ,\quad \text{where}\quad  \calJ^k=\sum_{h,s,v,a}
z^k_{h,s,v,a}\bigl(g^k_{h,s,v,a}\bigr)^2.
\end{equation*}

Summing over episodes telescopes the divergence terms. Since uniform initialization gives $D_{\mathcal H}(z^\star\Vert z^1)\le S(\log (|\mathcal V|)+\log A)$ and the divergence is nonnegative, we obtain
\begin{equation*}
\sum_{k=1}^K\langle z^k-z^\star,g^k\rangle
\le
\frac{S\log  (|\mathcal V|)+S\log A}{\lambda}
+\lambda\sum_{k=1}^K\calJ^k.
\end{equation*}

Unlike standard Exp3, the second-order term $\calJ^k$ cannot be bounded independently of the executed policy. In Exp3, the same action distribution is used for sampling and learning, so the sampling probabilities in the importance-weighted estimates cancel with those in the second-order term. Here, however, the learner executes the mixture policy $\mu_h^k$, determined by $q_h^k$, while OMD updates the joint weights $w^k(v)\pi_h^k(a\mid s,v)$. This mismatch, together with the candidate-dependent learning rates, prevents the same cancellation. Although setting $q_h^k=w^k$ would simplify the second-order term, it may leave a large discrepancy between the  loss predictions used at consecutive steps. Our key idea is therefore to choose $q^k$ to control both the second-order term and this discrepancy.

Given the above analysis and \pref{eq:reg_decom}, we have
\begin{equation}  \label{eq:expected_NRK_intermediate_step}
    \frac{\E [\NR_K]}{H} \leq \frac{S( \log  (|\mathcal V|)+\log A)}{\lambda}
+\sum_{k=1}^K \E \sbr{  L^k - \E^k \sbr{ \inner{z^k,g^k} } + \lambda \calJ^k }  .
\end{equation}

Hence, it remains to bound $\sum_{k=1}^K \E \sbr{  L^k - \E^k \sbr{ \inner{z^k,g^k} } + \lambda \calJ^k } $.

\subsubsection{Bounding $\sum_{k=1}^K \E \sbr{  L^k - \E^k \sbr{ \inner{z^k,g^k} } +\lambda \calJ^k }$.}\label{sec:bddrift}

Since both $L^k - \E^k \sbr{ \inner{z^k,g^k} }$ and $ \calJ^k$ depend on $q^k$, to bound $\sum_{k=1}^K \E \sbr{  L^k - \E^k \sbr{ \inner{z^k,g^k} } +\lambda \calJ^k }$, a natural idea is to minimize their sum over the feasible space of $q^k$. However, this minimization cannot be performed directly because both $L^k - \E^k \sbr{ \inner{z^k,g^k} }$ and $\calJ^k$ depend on the opponent's policy, which is unknown. 
Now, we show how to address the above issues. Specifically, rather than handling $\calJ^k$ or $\E^k[\calJ^k]$, we control an upper bound on $\E^k[\calJ^k]$.
Define the second-order cost as
\begin{equation}
\overline{\calJ}_{h}(z,s,q)
=\sum_v\frac{q_h(v\mid s)^2}{w(v)},
\qquad
\overline{\calJ}_{h}^k(s,q)=\overline{\calJ}_{h}(z^k,s,q).
\label{eq:bandit-cost}
\end{equation}

It then holds that $\E^k[\calJ^k] \leq \sum_{h,s}
\Prob^k(s_h^k = s)\overline{\calJ}_h^k(s,q^k)$.
To handle $L^k - \E^k \sbr{ \inner{z^k,g^k} }$,
we define $T^k = \{T^k_h\}_h$ as the
effective transition kernel induced by the opponent's policy $\nu^k$, given by
$T_h^k(s'\mid s,a)=\sum_{b\in\Bset_h}\nu_h^k(b\mid s)P_h(s'\mid s,a,b)$.
Conditional on $\mathcal F^k$, the committed policies and hence $T^k$ are fixed, although $T^k$ is unknown to the learner.

Let
$\mathcal T_{\mathrm{all}}=\prod_{h,s,a}\Delta(\mathcal S_{h+1})$
be a set containing every possible effective transition kernel.
For any fixed initial state $s_1\in\Sset_1$, joint variable $z\in\mathcal Z$
with induced $w$ and $\pi$,
mixing table $q$, and $T\in\mathcal T_{\mathrm{all}}$,
the state visitation probabilities satisfy
\begin{equation}
d_1^{z,q,T}(s)=\Indi\{s=s_1\},\qquad
d_{h+1}^{z,q,T}(s')=
\sum_{s,v,a}d_h^{z,q,T}(s)q_h(v\mid s)\pi_h(a\mid s,v)T_h(s'\mid s,a).\label{eq:defd}
\end{equation}
We further define
\begin{align}
\operatorname{Drift}(z,q,T)
& =
\sum_{h,s,v,a}
d_h^{z,q,T}(s)q_h(v\mid s)\pi_h(a\mid s,v)
\left(
v_h(s)-\sum_{s'}T_h(s'\mid s,a)v_{h+1}(s')
\right),\label{eq:def-drift}
\\\overline{\calJ}(z,q, T)  &  = \sum_{h,s} d^{z,q,T}_h(s) \overline{\calJ}_h(z,s,q).\label{eq:defB}
\end{align}

Based on these definitions, we have
\begin{align}
L^k - \E^k \sbr{ \inner{z^k,g^k} } & = \operatorname{Drift}(z^k,q^k,T^k)   \label{eq:drift}
\\ \E^k[\calJ^k] & \leq \sum_{h,s}\mathbb{P}^k(s_h^k = s)\overline{\calJ}_h^k(s,q^k)= \overline{\calJ}(z^k,q^k,T^k).\label{eq:Bk}
\end{align}
Consequently,
\[
\sum_{k=1}^K \E \sbr{
    L^k - \E^k \sbr{\inner{z^k,g^k}} + \lambda \calJ^k
}
\leq
\sum_{k=1}^K \E \sbr{
    \operatorname{Drift}(z^k,q^k,T^k)
    + \lambda \overline{\calJ}(z^k,q^k,T^k)
}.
\]
It therefore suffices to choose $q^k$ in each episode by solving
the following minimax problem:
\begin{align}
q^{k}\in\operatorname*{arg\,min}_{q\in \mathcal Q(z^k)}\Phi(z^k,q),\qquad
\Phi(z,q)=\sup_{T\in\mathcal T_{\mathrm{all}}}
 \left\{\operatorname{Drift}(z,q,T)+\lambda\overline{\calJ}(z,q,T)\right\},\label{eq:phiq}
\end{align}
where the feasible set is defined as %
\begin{align}
\mathcal Q(z)
=
\left\{
q:
\begin{array}{ll}
q_h(\cdot\mid s)\in\Delta(\mathcal V)
& \forall h,s,\\
q_1(\cdot\mid s)=w
& \forall s\in\mathcal S_1.
\\ \lambda q_h(v\mid s)\le w(v)& \forall h,s,v
\end{array}
\right\}. \label{eq:defqz}
\end{align}
The backward construction in the proof selects a common minimizer for all
$s_1\in\Sset_1$, so the mixing table can be chosen before the initial state is revealed.

In defining $\Phi$, we relax the actual effective transition kernel $T^k$ to the full class $\mathcal T_{\mathrm{all}}$. This relaxation reflects the inherent uncertainty about the current effective transition kernel. In particular, past trajectories are generated under unobserved opponent policies, which may change arbitrarily across episodes. In the unrestricted uninformed setting, any kernel $T\in\mathcal T_{\mathrm{all}}$ could be induced by a hidden opponent action that was never selected in previous episodes, leaving the distribution of historical observations unchanged. Thus, although the fixed underlying game may admit only a subset of $\mathcal T_{\mathrm{all}}$, the learner cannot, in general, rule out any particular kernel based on past observations alone.

As we show in \pref{lem:drift}, we have
\begin{align} 
     L^k - \E^k \sbr{ \inner{z^k,g^k} } +\lambda \E^k[\calJ^k] & \leq \operatorname{Drift}(z^k,q^k,T^k) + \lambda \overline{\calJ}(z^k,q^k,T^k)\nonumber
     \\ & \leq \Phi(z^k, q^{k})  =  \min_{q\in \mathcal Q(z^k)}\Phi(z^k,q) \leq 2S\lambda .\label{eq:residual_term_UB}
\end{align}

\begin{restatable}{lemma}{PhiUBsimpleversion} \label{lem:drift}
For any $0<\lambda\le 1/(2S)$ and $z\in\mathcal Z$ with induced weight $w$ and policy $\pi$ following \pref{eq:induce}, it holds that $\min_{q\in \mathcal{Q}(z)}\Phi(z,q) \leq  2S\lambda.$    
\end{restatable}

This implies $\sum_{k=1}^K \E \sbr{  L^k - \E^k \sbr{ \inner{z^k,g^k} } +\lambda \calJ^k }\leq 2SK\lambda$. 
The intuition behind the proof of \pref{lem:drift} is a tradeoff between controlling the drift and the second-order cost of learning. Without the second-order cost, the drift can always be made non-positive by placing the local weight $q_h(\cdot\mid s)$ on the candidate with the smallest prediction $v_h(s)$. Indeed, the resulting prediction at each state is no larger than the prediction inherited from the preceding step, ensuring a non-positive discrepancy regardless of the transition kernel. However, such an aggressive choice may concentrate on candidates with very small global weights $w(v)$, leading to a large second-order term due to importance weighting. We therefore choose $q_h$ to balance these two terms: including more candidates reduces the second-order term but may increase the drift error. This tradeoff ensures that the combined drift and second-order cost is at most $2S\lambda$, uniformly over all transition kernels. The proof of \pref{lem:drift} is provided in \pref{app:drift}

\subsubsection{Putting together}

Now, we plug \pref{eq:residual_term_UB} into \pref{eq:expected_NRK_intermediate_step} and use $\log  (|\mathcal V|) = \order  (S\log(HK))$ to show
\begin{equation*}
 \mathbb E[\NR_K/H]
 =   \otil \rbr{  S^2/\lambda
       +S\lambda K}.
\end{equation*}
Optimizing $\lambda$ gives
$\mathbb E[\NR_K]=\otil\big(HS^{3/2}\sqrt{K}\big)$.
We recall that this bound has no $A$ dependence because we have made the simplifying assumption on  full-information access to the one-step feedback,
and our final main result \pref{thm:main} not only addresses this issue (and naturally pays for a $\sqrt{A}$ factor), but also incorporates the following improvements: 
first, it is a high probability bound on the stronger regret notion $\ENR$;
second, it improves the dependence on $H$ and $S$ from $HS^{3/2}$ to $H^{3/2}S$.

\subsection{Related Work} Learning in Markov games has a long history, beginning with value-based
methods such as minimax-Q for two-player zero-sum games and Nash-Q for
general-sum stochastic games
\citep{littman1994markov,hu2003nash}.
More recent work has developed finite-sample guarantees for two-player
zero-sum Markov games, with a substantial line focusing on self-play, where
the learning procedure controls both players and can coordinate exploration
toward an equilibrium
\citep{bai2020provable,bai2020near,liu2021sharp,jin2024v}.
A complementary line studies online learning when a single learner interacts
with an arbitrary opponent. In the informed setting, where the opponent's
actions are observed, sublinear-regret guarantees have been established for
tabular average-reward stochastic games and episodic Markov games with
function approximation
\citep{wei2017online,xie2020learning}.
Our work studies the uninformed setting, where the opponent's actions are
hidden and the learner must learn both how to act and how to evaluate
continuation losses from its own trajectory feedback.

The closest line of work studies online learning against arbitrary opponents
when the opponent's actions are unobserved. \cite{tian2021online} introduce
the uninformed Markov-game setting and show that achieving sublinear external
regret generally requires an exponential dependence on the horizon. They
therefore turn to Nash-value regret and obtain an
$\widetilde O(K^{2/3})$ guarantee that is independent of the size of the
opponent's action space. \cite{appel2025regret} obtain an $\widetilde O(\sqrt{K})$ regret bound for robust online reinforcement learning, where the adversary can respond to the learner's realized action. While their setting allows a more powerful adversary, their regret is measured against a weaker turn-based value benchmark, rather than the empirical Nash-value benchmark considered here. Consequently, their regret guarantee cannot be directly applied to our setting.
More recently,
\cite{liu2026online} introduce empirical Nash-value regret ($\ENR$), a
stronger benchmark that recovers standard external regret against a fixed
opponent, and develop an algorithm that adapts to the opponent's
non-stationarity. Their guarantee achieves an $\widetilde O(\sqrt K)$ rate
for a fixed opponent while retaining an $\widetilde O(K^{2/3})$ rate in the
worst case. Our work closes this gap by establishing an
$\widetilde O(\sqrt K)$ worst-case guarantee for $\ENR$ against arbitrary
adaptive opponents.

Our approach is also related to exploration by optimization (ExO), which jointly selects exploration distributions and loss estimators by optimizing regret bounds that balance exploration costs, estimation bias, and the
stability of online updates~\cite{lattimore2020exploration,lattimore2021mirror}.
We follow a similar principle by choosing the local mixing weights $q^k$ to minimize the robust drift--cost objective in \pref{eq:phiq}. This objective combines the gap between local feedback and trajectory loss with a statistical cost, and we use the Bellman structure to bound it for all possible effective transition kernels.

\section{Algorithm and Main Result}
\label{sec:algorithm}

In this section, we present the complete algorithm
(\pref{alg:main}) and specify the key steps.
As described in \pref{subsec:technical-ingredients}, the algorithm follows the OMD framework, with  the joint domain \pref{eq:ti-joint-domain} as optimization space. Furthermore,  we adopt some modifications  to obtain
high-probability empirical Nash-value regret with better dependence on $S$. 
Throughout this section, we assume  $H\ge2$. The formal definition of $\mathcal{V}$ is given as 
\begin{equation}
\Vset=\left\{
v\in[0,1]^S:
\begin{aligned}
v_1(s)&=0,
&&\forall s\in\Sset_1,\\
v_h(s)&\in\left\{0,\frac{1}{HK},\ldots,
\frac{H-h+1}{H}\right\},
&&\forall h=2,\ldots,H,\ s\in\Sset_h
\end{aligned}
\right\}.
\label{eq:grid}
\end{equation}
We set $v_{H+1}(S_{H+1})=0$ for every $v\in\Vset$.

\paragraph{Refined regularizer.} In \pref{alg:main}, we slightly adjust the
scaling of the negative-entropy regularizer as follows:
\begin{equation}
 \mathcal H_{\mathrm{res}}(z)
 =H\sum_v w(v)\log w(v)
   +\sum_{h,s,v,a}z_{h,s,v,a}\log\frac{z_{h,s,v,a}}{w(v)}.
 \label{eq:omd-regularizer}
\end{equation}
Compared with \pref{eq:defent}, the only change is that
the coefficient of the candidate entropy is reduced from
$S$ to $H$. The intuition is that the candidate weights
are shared across all state steps,
 so their update must
account for the feedback accumulated across these steps.
However, in each episode, only $H$ state steps are visited
and contribute nonzero feedback, while all other steps
remain unchanged. Consequently, scaling the candidate
entropy by $H$ is sufficient to balance the contributions
of these active steps in the OMD regret analysis.
This allows us to preserve the OMD regret bound while
reducing the initial divergence from $S\log N$ to
$H\log N$. 

\paragraph{Implicit-exploration bandit feedback.} 
We estimate the remaining loss using the modified importance-weighted
estimator defined as
\begin{align}
\widehat Y_{h,s,v,a}^k=\frac{\Indi\{s_h^k=s,\ a_h^k=a\}}
        {\mu_h^k(a\mid s)+\eta_h^k(s,v)}
   \left(\frac{\ell_h^k}{H}+v_{h+1}(s_{h+1}^k)\right),\label{eq:omd-ix}
\end{align}
where $\eta_h^k(s,v) = \lambda q_h^k(v|s) /w^k(v)$. 
 As in implicit
exploration \citep{neu2015explore}, this modification introduces a
controlled downward bias while enabling exponential-moment bounds on
the estimation error. This control is crucial for obtaining
high-probability regret guarantees, rather than guarantees only in
expectation. At the terminal stage $h=H$, we set $v_{H+1}(\cdot)=0$
when computing $\widehat Y^k$.

\paragraph{Second-order cost for bandit feedback.}
To handle bandit feedback, we replace the quantity $Y_{h,s,v,a}^k$
used in the proof sketch of Section~\ref{subsec:technical-ingredients}
with the estimator $\widehat{Y}_{h,s,v,a}^k$ in \pref{eq:omd-ix}. 
Accordingly, for any $z\in\mathcal Z$ with induced weights $w$ and policies $\pi$, and any $q\in\mathcal Q(z)$,  with a
slight abuse of notation, we redefine the second-order cost in
\pref{eq:bandit-cost} as
\begin{align}
\overline{\calJ}_{h}(z,s,q)
=2\sum_v\frac{q_h(v\mid s)^2}{w(v)}
 \left(1+\sum_{a\in\mathcal A_h}
                 \frac{\pi_h(a\mid s,v)}{\mu_h^{z,q}(a\mid s)}\right),
\qquad
\overline{\calJ}_{h}^k(s,q)
=\overline{\calJ}_{h}(z^k,s,q),
\label{eq:bandit-cost-IX}
\end{align}
where $\mu_h^{z,q}$ is the mixture policy such that
$\mu_h^{z,q}(a\mid s)=\sum_v q_h(v\mid s)\pi_h(a\mid s,v)$.
We retain the definitions in \pref{eq:def-drift}, \pref{eq:defB}, and
\pref{eq:defqz}, replacing only the second-order cost in \pref{eq:bandit-cost}
with that in \pref{eq:bandit-cost-IX} and leaving all other terms unchanged.

The formal guarantee of \pref{alg:main} is stated below.

\begin{algorithm}[t]
\DontPrintSemicolon
\caption{A near-optimal algorithm for online learning in uninformed Markov games}
\label{alg:main}
\small
\textbf{Input:} episode $K$, 
and parameter $\lambda $ from \pref{thm:main}.

\textbf{Initialization:} the candidate grid in \pref{eq:grid} and the uniform joint
weights in \pref{eq:ti-mirror-step}.

\For{episode $k=1,\ldots,K$}{

Observe the initial state $s_1^k$.

Recover the candidate weights $w^k$ and action distributions $\pi^k$
from $z^k$ by \pref{eq:induce}.

Define $\operatorname{Drift}(z^k,q, T)$, $\overline{\calJ}(z^k,q,T)$ and $\mathcal{Q}(z^k)$ following \pref{eq:def-drift}, \pref{eq:defB}, \pref{eq:defqz} and \pref{eq:bandit-cost-IX}, with initial state $s_1 =s_1^k$.

 Compute  $q^k$ by solving
\begin{equation}
 q^k\in\operatorname*{arg\,min}_{q\in\mathcal Q(z^k)}
 \sup_{T\in\mathcal T_{\mathrm{all}}}
 \left\{\operatorname{Drift}(z^k,q,T)+\lambda\overline{\calJ}(z^k,q,T)\right\}. \label{eq:omd-planner}
\end{equation}

Set policy $\mu_h^k(a\mid s) = \mu^{z^k,q^k}_h(a\mid s)=\sum_{v\in\mathcal V}
q_h^k(v\mid s)\pi_h^k(a\mid s,v)$
for all $h\in[H]$, $s\in\Sset_h$, and $a\in\Aset_h$.

\For{$h=1,\ldots,H$}{The learner observes $s_h^k$, draws $a_h^k\sim\mu_h^k(\cdot\mid s_h^k)$,
and observes $\ell_h^k$ and $s_{h+1}^k$.}

Set $\eta_h^k(s,v)=\lambda q_h^k(v\mid s)/w^k(v)$
for all $h\in[H]$, $s\in\Sset_h$, and $v\in\mathcal V$.

For all $h\in[H]$, $s\in\Sset_h$, $v\in\mathcal V$,
and $a\in\Aset_h$, compute $\widehat Y_{h,s,v,a}^k$ by \pref{eq:omd-ix}, 
and $g_{h,s,v,a}^k$ as follows
\begin{align}
  g^k_{h,s,v,a}
&=\frac{  q^k_h(v|s)\Indi\{s_h^k=s\} }{ w^k(v) } 
\left(\widehat Y_{h,s,v,a}^k-v_h(s)\right).\label{eq:omd-feedback}
\end{align}
With $\mathcal{H}_{\mathrm{res}}(z)$ defined in \pref{eq:omd-regularizer} and $\mathcal Z$ defined in \pref{eq:ti-joint-domain}, update $z^{k+1}$ as
\begin{align}
    z^{k+1} = \argmin_{z\in \mathcal{Z}}\left\{\lambda \langle z, g^k \rangle  + D_{\mathcal{H}_{\mathrm{res}}}(z\|z^k)\right\}.\label{eq:omd_modified}
\end{align}

}
\end{algorithm}

\begin{theorem}
\label{thm:main}
For $\delta\in(0,1)$, let
$C_\delta=(H+1)\log N+2S\log A+(H+1)\log(2/\delta)$ and
$\lambda=\min\{[2(A+1)S]^{-1},\sqrt{C_\delta/[2(A+1)SK]}\}$.
\pref{alg:main} ensures that with probability at least $1-\delta$, $\ENR_K=\otil(H^{3/2}S\sqrt{AK})$.
\end{theorem}

The regret guarantee is independent of the opponent's
action-space size, even though the opponent's actions
are unobserved. 
The dependence on $K$ and $A$
is optimal up to logarithmic factors, while a gap
remains between the dependence on $H$ and $S$ and
the best known lower bounds (see \pref{sec:discussion}).
The full proof of \pref{thm:main} is postponed to \pref{app:omd-proof}.

\section{Discussion}\label{sec:discussion}
This paper establishes that worst-case
$\widetilde{O}(\sqrt{K})$ empirical Nash-value regret
is achievable in uninformed Markov games.
Nevertheless, two important questions remain open:

\vspace{-2ex}
\paragraph{Computational efficiency.}
Our algorithm explicitly maintains a class of
$N=\exp(\widetilde O(S))$ candidate value functions and requires
$O(SAN)$ storage. In each episode, it computes $q^k$ by backward
planning over state-step pairs. Assuming $N\gg AS$, the total
arithmetic complexity across $K$ episodes is
$\widetilde O\!\left(S^2 KN^{7/2}\log(KS)\right)$ (see \pref{lem:omd-planning}). Consequently,
the current implementation has exponential computational
and space costs. An important open question is whether
one can design a polynomial-time algorithm that retains
the same $\widetilde{O}(\sqrt{K})$ regret guarantee for uninformed Markov games.

\vspace{-2ex}
\paragraph{Closing the statistical gap.}
Although our regret bound achieves the optimal
dependence on $K$ and $A$, its dependence on $H$
and $S$ remains suboptimal. In particular, the
standard regret lower bound for episodic MDPs
implies $\Omega(H\sqrt{SAK})$~\cite{domingues2021episodic}, leaving a gap of $\widetilde{O}(\sqrt{HS})$ compared with our upper bound of $\widetilde{O}(H^{3/2}S\sqrt{AK})$.
It remains open whether this gap can be closed
by developing a sharper regret analysis or
establishing stronger lower bounds for
uninformed Markov games.

\section*{AI Use Statement.}

We used OpenAI's GPT-6 to assist with exploring and
refining algorithmic ideas, developing and refining
proof arguments, identifying connections to prior work,
and improving the exposition of the manuscript.
Its use materially influenced parts of the lower-bound
analysis in \pref{subsec:prior-limitations} and
\pref{app:epoch-v-learning-lb}, as well as the
algorithmic development and theoretical analysis in
\pref{subsec:algorithm-overview},
\pref{subsec:technical-ingredients},
\pref{sec:algorithm}, and \pref{app:omd-proof}.
The authors reviewed and verified all mathematical
claims, proofs, and references, and take full
responsibility for the correctness, originality,
and integrity of the paper.

\clearpage
\appendix
\section{Lower bound for epoch V-learning}
\label{app:epoch-v-learning-lb}

\LBVlearning*
\begin{proof}
We fix $K\ge1$ and $\eta\in(0,1/2]$.
Both constructions have a single initial state $x$, three actions at state $x$, and one action at each second-step state.
For $\eta\le K^{-1/3}$, a stationary opponent makes the learner pay for
identifying a worse action after each restart.
For $\eta>K^{-1/3}$, a single increase in the loss at a second-step state
remains absent from the first-step feedback until the current epoch ends.

\paragraph{Epoch schedule and feedback.}
The epoch lengths in Algorithm~1 of \cite{liu2026online} satisfy
$n_0=1$ and $n_j=\lceil(1+\eta)n_{j-1}\rceil$ for $j\ge1$.
Each state counts its own visits and starts a fresh local bandit at every
epoch boundary, with a uniform initial action distribution.
The new bandit uses only feedback from its current epoch.

In the loss convention, the second-step estimate $\widehat V_2^k(s)$
starts at zero and stays fixed during an epoch.
At the end of an epoch of length $n$, it becomes
$\max\{0,\overline\ell_2-\beta_n\}$, where $\overline\ell_2$ is the
average loss in that epoch and $\beta_n\ge0$.
This is the loss-form expression of Eq.~(3) of \cite{liu2026online}.
In particular, zero losses keep this estimate at zero.
With rewards $r_h=1-\ell_h$, the first-step bandit receives the
selected action and the scalar
$1-(\ell_1^k+\widehat V_2^k(s_2^k))/2$ within each epoch.
Thus, its policy depends on second-step observations through
$\widehat V_2^k$, which is updated for use in the next episode.

\paragraph{Case 1: frequent restarts, $\eta\le K^{-1/3}$.}
We first bound the number of visits in an epoch.
Since $n_j\ge n_{j-1}+1$, we have $n_j\ge j+1$.
If epoch $j$ starts within $K$ visits, then
$\sum_{i<j}n_i\le K-1$ and
$(j-1)(j+2)/2\le K-1$, so $j\le\sqrt{2K}$.
Summing $n_i-n_{i-1}\le\eta n_{i-1}+1$ over $i=1,\ldots,j$ gives
\[
 n_j\le1+\eta\left(1+\sum_{i<j}n_i\right)+j
 \le1+\eta K+\sqrt{2K}\le4K^{2/3}.
\]

Hence every observed epoch length is at most $4K^{2/3}$.
Let $\Delta=1/(48K^{1/3})$.
Every action at $x$ leads to the same second-step state, where the loss
is always zero.
The first-step loss matrix has the following three rows:
\[
\begin{array}{c|cccc}
 &b_1&b_2&b_3&b_4\\ \hline
 u&0&1&0&1\\
 v&0&0&1&1\\
 w&1&1&0&0
\end{array}
\]
For each $\theta\in\{1,2,3\}$, learner action $\theta$ is assigned row
$u$, and the other two actions are assigned rows $v,w$ in increasing
order.
The opponent independently samples from the same distribution
$\nu_\Delta=(1/4-\Delta/2,1/4+\Delta/2,1/4-\Delta/2,1/4+\Delta/2)$
in every episode.
The observed first-step loss is Bernoulli with mean $1/2+\Delta$ for
action $\theta$ and mean $1/2$ for either other action.
Mixing equally between rows $v,w$ gives loss $1/2$ against every opponent
action, while $\nu_\Delta$ gives expected loss at least $1/2$ against
every learner action. Therefore $V_1^\star(x)=1/2$.

Because the second-step estimate is always zero, the first-step bandit
learns from the first-step losses alone.
It is restarted at deterministic episode indices, since $x$ is visited
in every episode.
Let $m_j$ be the number of episodes observed in its $j$th epoch, including
the possibly shortened last epoch, and let $N_{\theta,j}$ count its
selections of action $\theta$ in that epoch.
We write $\mathbb P_{\theta,j}$ for the law of its action and loss sequence
in instance $\theta$.
Under the law $\mathbb P_{0,j}$, the same restarted bandit
instead receives independent Bernoulli losses of mean $1/2$ for every
action. The corresponding expectations are $\E_{\theta,j}$ and
$\E_{0,j}$.
Conditioning on the observations within the epoch, the action-selection
rule is the same under these laws; the conditional loss distributions
differ only when action $\theta$ is selected.
The chain rule for relative entropy therefore gives
\begin{align*}
 \KL(\mathbb P_{0,j}\Vert\mathbb P_{\theta,j})
 &=\E_{0,j}[N_{\theta,j}]
       \left(-\frac12\log(1-4\Delta^2)\right)\le4m_j\Delta^2\le\frac1{144}.
\end{align*}
Here $-\frac12\log(1-4\Delta^2)\le4\Delta^2$ since
$\Delta\le1/48$.
Pinsker's inequality gives
$\TV(\mathbb P_{0,j},\mathbb P_{\theta,j})\le\sqrt{1/288}<1/12$.
Since $0\le N_{\theta,j}\le m_j$, it follows that
$\E_{\theta,j}[N_{\theta,j}]\ge\E_{0,j}[N_{\theta,j}]-m_j/12$.
As the three action counts sum to $m_j$,
\[
 \frac13\sum_{\theta=1}^3\E_{\theta,j}[N_{\theta,j}]
 \ge\frac{m_j}{3}-\frac{m_j}{12}=\frac{m_j}{4}.
\]
In instance $\theta$, the expected regret is
$\E_\theta[\NR_K]=\Delta\sum_j\E_{\theta,j}[N_{\theta,j}]$,
because the conditional expected excess loss is $\Delta$ times the
probability of selecting action $\theta$.
Summing over epochs, using $\sum_jm_j=K$, yields
\[
 \frac13\sum_{\theta=1}^3\E_\theta[\NR_K]
 \ge\frac{\Delta}{4}\sum_jm_j
 =\frac{K^{2/3}}{192}.
\]
Thus at least one fixed instance $\theta$ has the claimed regret.

\paragraph{Case 2: delayed response, $\eta>K^{-1/3}$.}
There are two second-step states, a safe state $g$ and a risky state $z$.
All first-step losses are zero.
One action (of the learner) $a^\dagger$ at state $x$ leads to state $z$, and the other two lead
to state $g$.
The opponent has two actions $b_0,b_1$ at the second step, with
$\ell_2(g,b)=0$ for both $b$,
$\ell_2(z,b_0)=0$, and $\ell_2(z,b_1)=1$.
For a switch time $\tau\in\{1,\ldots,K\}$, the opponent plays $b_0$ at
state $z$ before episode $\tau$ and $b_1$ there from episode $\tau$ onward.
Its policy elsewhere is fixed.
A safe action guarantees zero total loss, so $V_1^\star(x)=0$.

We determine $a^\dagger$ using a reference run in which the opponent
always plays $b_0$.
Both second-step estimates remain zero, so the feedback at state $x$ is always $1$.
The distribution of the first-step action sequence is therefore the
same for every assignment of the action leading to state $z$.
Since the three action counts sum to $K$, an action $a^\dagger$ has
expected count at least $K/3$ in this reference run.
With this assignment fixed, let $M$ be the number of reference visits to
$z$; then $\E_{\mathrm{ref}}[M]\ge K/3$.

For a fixed $\tau$, the reference run and the switched run use the same
randomness in their local bandits and action draws.
They agree before $\tau$.
At the first visit to $z$ at or after $\tau$, its current estimate is
still zero, and it remains zero until that local epoch ends.
The estimate at $g$ stays zero throughout.
Until this epoch ends, the first-step bandit receives the same feedback
in both runs and hence selects the same actions.
Let $D_\tau$ be the number of reference visits to $z$ at or after
$\tau$ in this epoch, up to episode $K$, with $D_\tau=0$ if there is no
such visit.
All these visits also occur in the switched run and incur unit loss.
The estimate changes only after the last visit of a completed epoch.
Since the Nash value and the first-step losses are zero,
\[
 \E_\tau[\NR_K]
 =\E_\tau\!\left[\sum_{k=1}^K\ell_2^k\right]
 \ge\E_{\mathrm{ref}}[D_\tau].
\]

We average this inequality over the $K$ deterministic switch times.
For a reference trajectory, let $m_1,\ldots,m_{N_{\mathrm{ep}}}$ be the observed lengths
of the epochs at $z$, so $\sum_jm_j=M$.
If $\tau$ is the episode of the $i$th visit in epoch $j$, then
$D_\tau=m_j-i+1$.
These visits give distinct switch times, and every $D_\tau$ is
nonnegative. Thus
\[
 \sum_{\tau=1}^K D_\tau
 \ge\sum_{j=1}^{N_{\mathrm{ep}}}\sum_{i=1}^{m_j}(m_j-i+1)
 =\sum_{j=1}^{N_{\mathrm{ep}}}\frac{m_j(m_j+1)}2
 \ge\frac12\sum_{j=1}^{N_{\mathrm{ep}}}m_j^2.
\]

To bound this square sum, let $r$ be the number of completed epochs
at $z$, and let $d$ be the number of visits in the unfinished epoch,
with $d=0$ if there is none.
Then $M=\sum_{j=1}^rn_j+d$ and
$\sum_{j=1}^{N_{\mathrm{ep}}}m_j^2=\sum_{j=1}^rn_j^2+d^2$.
The geometric growth of the completed epoch lengths implies
$\sum_{i<j}n_i\le n_j\sum_{t\ge1}(1+\eta)^{-t}=n_j/\eta$.
Expanding the square gives
\[
 \left(\sum_{j=1}^rn_j\right)^2
 =\sum_{j=1}^r\left(n_j^2+2n_j\sum_{i<j}n_i\right)
 \le\left(1+\frac2\eta\right)\sum_{j=1}^rn_j^2.
\]
Using $(a+b)^2\le2a^2+2b^2$ and $\eta\le1/2$, we obtain
\begin{align*}
 M^2
 &\le2\left(\sum_{j=1}^rn_j\right)^2+2d^2\le2\left(1+\frac2\eta\right)\sum_{j=1}^rn_j^2+2d^2
 \le\frac5\eta\sum_{j=1}^{N_{\mathrm{ep}}}m_j^2.
\end{align*}
Combining the preceding bounds and
$\E_{\mathrm{ref}}[M^2]\ge(\E_{\mathrm{ref}}[M])^2\ge K^2/9$
yields
\begin{align*}
 \frac1K\sum_{\tau=1}^K\E_\tau[\NR_K]
 &\ge\frac1{2K}\E_{\mathrm{ref}}\!\left[\sum_{j=1}^{N_{\mathrm{ep}}}m_j^2\right]\ge\frac\eta{10K}\E_{\mathrm{ref}}[M^2]
 \ge\frac{\eta K}{90}
 >\frac{K^{2/3}}{90}.
\end{align*}
Hence some deterministic switch time $\tau$ gives this lower bound.
Together with Case~1, this proves
$\E[\NR_K]\ge K^{2/3}/192$ for every fixed $\eta\in(0,1/2]$.
\end{proof}

\section{ Omitted Details in \pref{sec:algorithm} }
\label{app:omd-proof}

For shorthand, we define
\[
 Y_{h,s,v,a}^k
 =\sum_b\nu_h^k(b\mid s)
   \left[\frac{\ell_h(s,a,b)}H+P_hv_{h+1}(s,a,b)\right].
\]

\subsection{Supporting Lemmas}
\subsubsection{OMD Analysis}
The regularizer $\mathcal H_{\mathrm{res}}$ in \pref{eq:omd-regularizer} has divergence
\[
 D_{\mathcal H_{\mathrm{res}}}(z'\Vert z)
 =H\operatorname{KL}(w'\Vert w)
   +\sum_{h,s,v}w'(v)
     \operatorname{KL}\bigl(\pi'_h(\cdot\mid s,v)
                           \Vert\pi_h(\cdot\mid s,v)\bigr),
\]
where terms with $w'(v)=0$ contribute zero. For a candidate $v$ and a mixed
Markov policy $x$, let
$z^{v,x}_{h,s,v',a}=\Indi\{v'=v\}x_h(a\mid s)$.
Uniform initialization gives
$D_{\mathcal H_{\mathrm{res}}}(z^{v,x}\Vert z^1)\le H\log N+S\log A$.
For shorthand, we define
\[
M_{h,s,v}^k=\sum_a\pi_h^k(a\mid s,v)e^{-\lambda g_{h,s,v,a}^k},\quad  Q^k=\sum_vw^k(v)\rbr{\prod_{h,s}M_{h,s,v}^k}^{1/H}.
\]

\begin{lemma}
\label{lem:omd-update}
For every $z\in\mathcal Z$ and every episode $k$,
\begin{equation}
 D_{\mathcal H_{\mathrm{res}}}(z\Vert z^{k+1})-D_{\mathcal H_{\mathrm{res}}}(z\Vert z^k)
 =H\log Q^k+\lambda\langle z,g^k\rangle.
 \label{eq:omd-exact-identity}
\end{equation}
Moreover,
\begin{equation}
 Q^k\le 1+\frac1H\sum_{h=1}^H\sum_v
             w^k(v)(M_{h,s_h^k,v}^k-1).
 \label{eq:omd-normalizer-average}
\end{equation}
\end{lemma}

\begin{proof}
For a feasible $z'=w'\pi'$, the OMD objective in
\pref{eq:omd_modified}, using the divergence above, equals
\begin{align*}
 H\operatorname{KL}(w'\Vert w^k)
 +\sum_{h,s,v}w'(v)\left[
 \lambda\langle\pi'_h(\cdot\mid s,v),g_{h,s,v}^k\rangle
 +\operatorname{KL}\bigl(\pi'_h(\cdot\mid s,v)
                         \Vert\pi_h^k(\cdot\mid s,v)\bigr)\right].
\end{align*}
For fixed $w'$, each action distribution $\pi'_h(\cdot\mid s,v)$ can be optimized independently.
By the definition of KL divergence,
\begin{align*}
&\lambda\langle\pi'_h(\cdot\mid s,v),g_{h,s,v}^k\rangle
+\operatorname{KL}\bigl(\pi'_h(\cdot\mid s,v)
\Vert\pi_h^k(\cdot\mid s,v)\bigr)\\
&=\sum_a\pi'_h(a\mid s,v)
\left(\lambda g_{h,s,v,a}^k
+\log\frac{\pi'_h(a\mid s,v)}{\pi_h^k(a\mid s,v)}\right)\\
&=\sum_a\pi'_h(a\mid s,v)
\log\frac{\pi'_h(a\mid s,v)}
{\pi_h^k(a\mid s,v)e^{-\lambda g_{h,s,v,a}^k}}\\
&=\sum_a\pi'_h(a\mid s,v)
\log\frac{\pi'_h(a\mid s,v)}
{M_{h,s,v}^k\pi_h^{k+1}(a\mid s,v)}\\
&=\operatorname{KL}\bigl(\pi'_h(\cdot\mid s,v)
\Vert\pi_h^{k+1}(\cdot\mid s,v)\bigr)
-\log M_{h,s,v}^k,
\end{align*}
where the penultimate equality follows from
$\pi_h^{k+1}(a\mid s,v)
=\pi_h^k(a\mid s,v)e^{-\lambda g_{h,s,v,a}^k}/M_{h,s,v}^k$,
and the last equality uses $\sum_a\pi'_h(a\mid s,v)=1$.
Nonnegativity of KL shows that this is the minimizing distribution,
with minimum value $-\log M_{h,s,v}^k$.
Consequently, the remaining objective is
\[
H\operatorname{KL}(w'\Vert w^k)
-\sum_vw'(v)\sum_{h,s}\log M_{h,s,v}^k =H\operatorname{KL}(w'\Vert w^{k+1})-H\log Q^k,
\]
where the equality follows from the definition of $Q^k$, and 
\[
w^{k+1}(v)=w^k(v)\exp(H^{-1}\sum_{h,s}\log M_{h,s,v}^k)/Q^k.
\]
Thus the displayed updates minimize the joint OMD objective and preserve
positivity.

\needspace{9\baselineskip}
For an arbitrary comparison point $z=w\pi$, the two updates give
$\log(w^k(v)/w^{k+1}(v))
=\log Q^k-H^{-1}\sum_{h,s}\log M_{h,s,v}^k$ and
$\log(\pi_h^k(a\mid s,v)/\pi_h^{k+1}(a\mid s,v))
=\lambda g_{h,s,v,a}^k+\log M_{h,s,v}^k$.
Substituting into the divergence and using
$\sum_a z_{h,s,v,a}=w(v)$ yields
\begin{align*}
 D_{\mathcal H_{\mathrm{res}}}(z\Vert z^{k+1})-D_{\mathcal H_{\mathrm{res}}}(z\Vert z^k)
 &=H\log Q^k-\sum_{h,s,v}w(v)\log M_{h,s,v}^k
       +\lambda\langle z,g^k\rangle
       +\sum_{h,s,v}w(v)\log M_{h,s,v}^k\\
 &=H\log Q^k+\lambda\langle z,g^k\rangle,
\end{align*}
which proves \pref{eq:omd-exact-identity}.
Finally, $g_{h,s,v,a}^k=0$ at every unvisited state, so
$M_{h,s,v}^k=1$ there and
$\prod_{h,s}M_{h,s,v}^k=\prod_{h=1}^H M_{h,s_h^k,v}^k$.
Applying the arithmetic--geometric mean inequality separately for each
$v$ gives
$(\prod_{h=1}^H M_{h,s_h^k,v}^k)^{1/H}
\le H^{-1}\sum_{h=1}^H M_{h,s_h^k,v}^k$.
Multiplying by $w^k(v)$, summing over $v$, and using
$\sum_vw^k(v)=1$ gives \pref{eq:omd-normalizer-average}.
\end{proof}

\begin{lemma}
\label{lem:omd-local}
For every episode $k$ and step $h$, with $s=s_h^k$, we have
\begin{align*}
\E_h^k\left[\sum_v w^k(v)(M_{h,s,v}^k-1)\right]\le -\lambda\sum_v q_h^k(v\mid s)
\left(\sum_a\pi_h^k(a\mid s,v)Y_{h,s,v,a}^k-v_h(s)\right)
+\lambda^2\overline{\calJ}_h^k(s,q^k).
\end{align*}
\end{lemma}

\begin{proof}
At the visited state $s=s_h^k$, the IX estimator in
\pref{eq:omd-ix} satisfies
\[
\E_h^k[\widehat Y_{h,s,v,a}^k]
=\frac{\mu_h^k(a\mid s)Y_{h,s,v,a}^k}
{\mu_h^k(a\mid s)+\eta_h^k(s,v)}.
\]

Since $0\le Y_{h,s,v,a}^k\le1$ and $\eta_h^k(s,v)=\lambda q_h^k(v\mid s)/w^k(v)$,
the resulting bias is bounded by
\begin{align*}
&\sum_{v,a}q_h^k(v\mid s)\pi_h^k(a\mid s,v)
\left(Y_{h,s,v,a}^k-\E_h^k[\widehat Y_{h,s,v,a}^k]\right)\\
&=\sum_{v,a}q_h^k(v\mid s)\pi_h^k(a\mid s,v)
\frac{\eta_h^k(s,v)Y_{h,s,v,a}^k}
{\mu_h^k(a\mid s)+\eta_h^k(s,v)}\\
&\le\lambda\sum_{v,a}
\frac{q_h^k(v\mid s)^2}{w^k(v)}
\frac{\pi_h^k(a\mid s,v)}{\mu_h^k(a\mid s)}\\
&\le\frac{\lambda}{2}\overline{\calJ}_h^k(s,q^k), \numberthis{} \label{eq:YminusEY_UB_lambdaB}
\end{align*}
where the last inequality follows from \pref{eq:bandit-cost-IX}.

Since $\ell_h^k/H+v_{h+1}(s_{h+1}^k)$ and $v_h(s)$
both belong to $[0,1]$, we have
\begin{align*}
\E_h^k\left[
(\widehat Y_{h,s,v,a}^k-v_h(s))^2\right]
&\le \E_h^k[(\widehat Y_{h,s,v,a}^k)^2]+v_h(s)^2\\
&\le\frac{\mu_h^k(a\mid s)}
{(\mu_h^k(a\mid s)+\eta_h^k(s,v))^2}+1\\
&\le\frac{1}{\mu_h^k(a\mid s)}+1.
\end{align*}
Multiplying by
$q_h^k(v\mid s)^2\pi_h^k(a\mid s,v)/w^k(v)$
and summing over $v,a$ gives
\begin{align} \label{eq:wpiEg_UB_B}
&\sum_{v,a}w^k(v)\pi_h^k(a\mid s,v)
\E_h^k[(g_{h,s,v,a}^k)^2]
\le \frac{1}{2}\overline{\calJ}_h^k(s,q^k).
\end{align}

Finally, the constraint in \pref{eq:omd-planner} ensures
$\lambda g_{h,s,v,a}^k
\ge-\eta_h^k(s,v)v_h(s)\ge-1$.
Applying $e^{-t}\le1-t+t^2$ for $t\ge-1$ to the definition
of $M_{h,s,v}^k$, we obtain
\begin{align*}
&\E_h^k\left[\sum_vw^k(v)(M_{h,s,v}^k-1)\right]\\
&\le-\lambda\sum_{v,a}w^k(v)\pi_h^k(a\mid s,v)
\E_h^k[g_{h,s,v,a}^k]
+\lambda^2\sum_{v,a}w^k(v)\pi_h^k(a\mid s,v)
\E_h^k[(g_{h,s,v,a}^k)^2]\\
&=-\lambda\sum_{v,a}q_h^k(v\mid s)\pi_h^k(a\mid s,v)
\left(\E_h^k[\widehat Y_{h,s,v,a}^k]-v_h(s)\right)
+\lambda^2\sum_{v,a}w^k(v)\pi_h^k(a\mid s,v)
\E_h^k[(g_{h,s,v,a}^k)^2]\\
&=-\lambda\sum_{v,a}q_h^k(v\mid s)\pi_h^k(a\mid s,v)
\left(Y_{h,s,v,a}^k-v_h(s)\right)\\
&\quad+\lambda\sum_{v,a}q_h^k(v\mid s)\pi_h^k(a\mid s,v)
\left(Y_{h,s,v,a}^k-\E_h^k[\widehat Y_{h,s,v,a}^k]\right)\\
&\quad+\lambda^2\sum_{v,a}w^k(v)\pi_h^k(a\mid s,v)
\E_h^k[(g_{h,s,v,a}^k)^2]\\
&\le-\lambda\sum_vq_h^k(v\mid s)
\left(\sum_a\pi_h^k(a\mid s,v)Y_{h,s,v,a}^k-v_h(s)\right)
+\lambda^2\overline{\calJ}_h^k(s,q^k),
\end{align*}
where the last inequality uses \pref{eq:wpiEg_UB_B} and \pref{eq:YminusEY_UB_lambdaB}.
\end{proof}

\begin{lemma}
\label{lem:omd-learner}
With probability at least $1-\delta/2$, for every candidate
$v\in\calV$ and every mixed Markovian policy $x$,
\begin{equation}
\sum_{k=1}^K\bigl(L^k-\langle z^{v,x},g^k\rangle\bigr)
\le
\frac{H\log|\calV|+S\log A+H\log(2/\delta)}{\lambda}
+2(A+1)S\lambda K.
\label{eq:omd-learner-bound}
\end{equation}
\end{lemma}

\begin{proof}
The actual effective transition kernel $T^k$ belongs to
$\mathcal T_{\mathrm{all}}$.
By \pref{eq:def-drift}, the sum of the true residuals in
\pref{lem:omd-local}, averaged over state visits, equals
$L^k-\operatorname{Drift}(z^k,q^k,T^k)$.
Taking conditional expectation in
\pref{eq:omd-normalizer-average}, followed by
\pref{lem:omd-local} and \pref{lem:omd-planning}, gives
\begin{align*}
\E^k[Q^k]
&\le 1+\frac{\lambda}{H}
\left[-L^k+\operatorname{Drift}(z^k,q^k,T^k)
+\lambda\overline{\calJ}(z^k,q^k,T^k)\right]\\
&\le\exp\left(
\frac{\lambda}{H}
\bigl(2(A+1)S\lambda-L^k\bigr)
\right).
\end{align*}

Since $L^k$ is measurable with respect to $\mathcal F^k$,
the preceding inequality implies
\[
\E^k\left[
Q^k\exp\left(
\frac{\lambda}{H}
\bigl(L^k-2(A+1)S\lambda\bigr)
\right)\right]\le1.
\]
Moreover, the quantities from episodes $1,\ldots,k-1$
are measurable with respect to $\mathcal F^k$.
Applying the tower property repeatedly, starting from
episode $K$, yields
\begin{align*}
&\E\left[
\prod_{k=1}^K\Bigl[Q^k
\exp\left(
\frac{\lambda}{H}
\bigl(L^k-2(A+1)S\lambda\bigr)
\right)\Bigr]\right]\\
&=\E\Bigg[
\prod_{k=1}^{K-1}\Bigl[Q^k
\exp\left(
\frac{\lambda}{H}
\bigl(L^k-2(A+1)S\lambda\bigr)
\right)\Bigr] \cdot
\E^K\left[
Q^K\exp\left(
\frac{\lambda}{H}
\bigl(L^K-2(A+1)S\lambda\bigr)
\right)\right]\Bigg]\\
&\le\E\left[
\prod_{k=1}^{K-1}\Bigl[Q^k
\exp\left(
\frac{\lambda}{H}
\bigl(L^k-2(A+1)S\lambda\bigr)
\right)\Bigr]\right]\\
&\le\cdots\le1.
\end{align*}
Here, no independence across episodes is required.
By Markov's inequality,
\[
\P\left(
\prod_{k=1}^K\Bigl[Q^k
\exp\left(
\frac{\lambda}{H}
\bigl(L^k-2(A+1)S\lambda\bigr)
\right)\Bigr]
\ge\frac2\delta
\right)\le\frac\delta2.
\]
Therefore, with probability at least $1-\delta/2$,
taking logarithms and rearranging gives
\begin{align*}
\sum_{k=1}^K L^k
+\frac H\lambda\sum_{k=1}^K\log Q^k
\le 2(A+1)S\lambda K
+\frac H\lambda\log\frac2\delta.
\end{align*}

For any pair $(v,x)$, summing
\pref{eq:omd-exact-identity} over episodes gives
\begin{align*}
&D_{\mathcal H_{\mathrm{res}}}(z^{v,x}\Vert z^{K+1})
-D_{\mathcal H_{\mathrm{res}}}(z^{v,x}\Vert z^1) =H\sum_{k=1}^K\log Q^k
+\lambda\sum_{k=1}^K\langle z^{v,x},g^k\rangle.
\end{align*}
Using nonnegativity of the terminal divergence and
$D_{\mathcal H_{\mathrm{res}}}(z^{v,x}\Vert z^1)
\le H\log|\calV|+S\log A$, we obtain
\begin{align*}
-\frac H\lambda\sum_{k=1}^K\log Q^k
&\le\frac{D_{\mathcal H_{\mathrm{res}}}(z^{v,x}\Vert z^1)}{\lambda}
+\sum_{k=1}^K\langle z^{v,x},g^k\rangle \le\frac{H\log|\calV|+S\log A}{\lambda}
+\sum_{k=1}^K\langle z^{v,x},g^k\rangle.
\end{align*}
Combining this inequality with the preceding
high-probability bound proves the claim.
Since the high-probability event does not depend on
$(v,x)$ and the OMD identity holds for every comparator
on each realized trajectory, the bound holds
simultaneously for all comparison pairs.
\end{proof}

\begin{lemma}
\label{lem:omd-uniform-ix}
With probability at least $1-\delta/2$, for every
candidate $v\in\calV$ and every mixed Markov policy $x$,
\begin{equation}
\begin{aligned}
&\sum_{k=1}^K\sum_{h=1}^H
\frac{q_h^k(v\mid s_h^k)}{w^k(v)}
\sum_a x_h(a\mid s_h^k)
\bigl(\widehat Y_{h,s_h^k,v,a}^k
-Y_{h,s_h^k,v,a}^k\bigr)\le
\frac{\log|\calV|+S\log A+\log(2/\delta)}{\lambda}.
\end{aligned}
\label{eq:omd-uniform-ix}
\end{equation}
\end{lemma}

\begin{proof}
Fix an episode $k$, step $h$, candidate $v$, and action $a$.
At the visited state $s=s_h^k$, write
$p=\mu_h^k(a\mid s)>0$ and
$\eta=\eta_h^k(s,v)\ge0$ for shorthand.
For any $u\in[0,1]$, the inequality
$\log(1+t)\ge t/(1+t)$ for $t\ge0$ gives
\[
\frac{\eta u}{p+\eta}
\le\frac{\eta u}{p+\eta u}
\le\log\left(1+\frac{\eta u}{p}\right).
\]
Consequently,
\[
\exp\left(\frac{\eta u}{p+\eta}\right)
\le1+\frac{\eta u}{p}.
\]

By \pref{eq:omd-ix}, the estimator is zero when
$a_h^k\ne a$, and equals
$(\ell_h^k/H+v_{h+1}(s_{h+1}^k))/(p+\eta)$
when $a_h^k=a$.
Since $\ell_h^k/H+v_{h+1}(s_{h+1}^k)\in[0,1]$,
we obtain
\begin{align*}
&\E_h^k\left[
\exp\left(\eta\widehat Y_{h,s,v,a}^k\right)\right]\\
&=1-p+p\E_h^k\left[
\exp\left(
\frac{\eta}{p+\eta}
\left(\frac{\ell_h^k}{H}
+v_{h+1}(s_{h+1}^k)\right)
\right)\,\middle|\,a_h^k=a\right]\\
&\le1-p+p\E_h^k\left[
1+\frac{\eta}{p}
\left(\frac{\ell_h^k}{H}
+v_{h+1}(s_{h+1}^k)\right)
\,\middle|\,a_h^k=a\right]\\
&=1+\eta Y_{h,s,v,a}^k,
\end{align*}
where the last equality follows from the definition of
$Y_{h,s,v,a}^k$.
Since $Y_{h,s,v,a}^k$ is measurable with respect to
$\mathcal F_h^k$ and $1+t\le e^t$, this implies
\begin{align*}
\E_h^k\left[
\exp\left(
\eta(\widehat Y_{h,s,v,a}^k-Y_{h,s,v,a}^k)
\right)\right]\le(1+\eta Y_{h,s,v,a}^k)
e^{-\eta Y_{h,s,v,a}^k}
\le1.
\end{align*}

Now fix a candidate $v\in\calV$ and a deterministic
Markov policy $f$.
At every visited state, the preceding inequality
applies to the action $a=f_h(s_h^k)$.
Since
$\eta_h^k(s,v)=\lambda q_h^k(v\mid s)/w^k(v)$,
we have
\begin{align*}
\E_h^k\Bigg[
\exp\Bigg(
\lambda\frac{q_h^k(v\mid s_h^k)}{w^k(v)}
\bigl(
\widehat Y_{h,s_h^k,v,f_h(s_h^k)}^k
-Y_{h,s_h^k,v,f_h(s_h^k)}^k
\bigr)
\Bigg)\Bigg]\le1.
\end{align*}

The candidate weights and mixing weights at step $h$
are determined by the history before the action is
sampled.
Thus, conditioning successively on
$\mathcal F_H^K,\ldots,\mathcal F_1^K$ and then
on the corresponding histories of earlier episodes
gives
\begin{align*}
\E\Bigg[
\exp\Bigg(
\lambda\sum_{k=1}^K\sum_{h=1}^H
\frac{q_h^k(v\mid s_h^k)}{w^k(v)}
\bigl(
\widehat Y_{h,s_h^k,v,f_h(s_h^k)}^k
-Y_{h,s_h^k,v,f_h(s_h^k)}^k
\bigr)
\Bigg)\Bigg]\le1.
\end{align*}

Applying Markov's inequality, for this fixed pair
$(v,f)$ we obtain
\begin{align*}
&\P\Bigg(
\sum_{k=1}^K\sum_{h=1}^H
\frac{q_h^k(v\mid s_h^k)}{w^k(v)}
\bigl(
\widehat Y_{h,s_h^k,v,f_h(s_h^k)}^k
-Y_{h,s_h^k,v,f_h(s_h^k)}^k
\bigr)>
\frac{\log|\calV|+S\log A+\log(2/\delta)}
{\lambda}
\Bigg)
\le\frac{\delta}{2|\calV|A^S}.
\end{align*}

There are $N= |\calV|$ candidates and at most $A^S$
deterministic Markov policies, since an action is
specified at each of the $S$ states.
A union bound therefore shows that, with probability
at least $1-\delta/2$, the desired inequality holds
simultaneously for every candidate $v$ and every
deterministic Markov policy $f$.
Finally, every mixed Markov policy $x$ can be written
as a convex combination of deterministic Markov
policies. For each realized history, the left-hand
side of \pref{eq:omd-uniform-ix} is linear in
$x_h(a\mid s)$.
It is therefore a convex combination of the
corresponding expressions for deterministic policies.
Since each of these expressions satisfies the same
upper bound on the event established above, the
inequality also holds for every mixed Markov policy
$x$, completing the proof.
\end{proof}

\subsubsection{Drift--Cost Bound}\label{app:drift}

We first present the bound of general drift-cost bound, and then apply it to prove \pref{lem:omd-planning} and \pref{lem:drift}.

\begin{lemma}[General drift--cost bound]
\label{lem:general-drift-cost}
Let
$0<\lambda\le 1/(2S)$.
Fix $z\in\mathcal Z$ with induced weights $w$ and policies $\pi$.
Assume that every candidate $v\in\mathcal V$ satisfies
$v_1=v_{H+1}=0$ and $0\le v_h(s)\le 1$.
For each $(h,s)$, let $R_h(s, p)$ be a nonnegative local cost
defined on distribution $p\in\Delta(\mathcal{V})$ supported on $\operatorname{supp}(w)$.
Define
\begin{align}
\Phi_R(z,q) =\sup_{T\in \mathcal{T}_{\mathrm{all}}}\left\{\operatorname{Drift}(z,q,T)
+\sum_{h,s}d_h^{z,q,T}(s)
R_h(s,q_h(\cdot|s))\right\}.
\end{align}
We write $m_G=\sum_{v\in G}w(v)$ and $p^G(v)=w(v)\Indi\{v\in G\}/m_G$.
Suppose that there exists $c\ge 2\lambda$ such that, for every
nonempty subset $G\subseteq\operatorname{supp}(w)$, we have
$R_h(s,p^G)\le c/m_G$ for all $h\in[H]$ and $s\in\mathcal S_h$.
Then 
$\min_{q\in \mathcal{Q}(z)}\Phi_R(z,q)\le cS$.
\end{lemma}

\begin{proof}
Let $m_G$ and $p^G$ be as in the statement, and define
$\mathcal P_\lambda(w)
=\{p\in \Delta(\mathcal{V}): \lambda p(v) \leq w(v), \forall v\in \mathcal{V}\}$.
We optimize over distributions satisfying $q_1(\cdot\mid s)=w$
and $q_h(\cdot\mid s)\in\mathcal P_\lambda(w)$ for all $h\ge2$.

\paragraph{Bellman optimization.}
For a distribution $p\in \Delta(\mathcal{V})$ and a function $f$ on $\mathcal S_{h+1}$,
define
\begin{align*}
\mathcal F_h(s,p;f)
&=
R_h(s,p)+\sum_v p(v)v_h(s) +
\sum_{a\in\mathcal A_h}
\max_{s'\in\mathcal S_{h+1}}
\sum_v p(v)\pi_h(a\mid s,v)
\bigl(f(s')-v_{h+1}(s')\bigr).
\end{align*}
Set $\Psi_{H+1}=0$.
For $h=H,\ldots,2$, define
\begin{align}
\Psi_h(s)
=
\min_{p\in\mathcal P_\lambda(w)}
\mathcal F_h(s,p;\Psi_{h+1}), \quad\qquad 
q_h^\star(\cdot\mid s)
\in
\operatorname*{arg\,min}_{p\in\mathcal P_\lambda(w)}
\mathcal F_h(s,p;\Psi_{h+1}).\label{eq:optr}
\end{align}
At the initial step, set $q_1^\star(\cdot\mid s)=w$ and
$\Psi_1(s)=\mathcal F_1(s,w;\Psi_2)$.

The maximization over successor states in $\mathcal F_h$ is exact
because the transition rows in $\mathcal T_{\mathrm{all}}$
can be chosen independently for each $(h,s,a)$.
Moreover, $\mathcal F_h(s,p;f)$ is nondecreasing in $f$.
Consequently, backward induction shows that $q^\star$ minimizes
$\Phi_R(z,q)$ over the specified family and that
$\Phi_R(z,q^\star)=\Psi_1(s_1)$.
It remains to bound this optimal value.

\paragraph{Bounding the optimal values.}
Define
$\theta_h(v)=\max_{s'\in\mathcal S_{h+1}}
[\Psi_{h+1}(s')-v_{h+1}(s')]_+$.
We prove by backward induction that
$\sum_v w(v)\theta_h(v)
\le c\sum_{j=h+1}^H|\mathcal S_j|$.
The claim holds for $h=H$, since $\Psi_{H+1}=v_{H+1}=0$.

Suppose the claim holds at step $h\ge2$.
Fix $s\in\mathcal S_h$ and write
$a_v=v_h(s)+\theta_h(v)$.
Introduce the comparison value
$U_h(s)=
\min_{\varnothing\ne G\subseteq\operatorname{supp}(w)}
\{\sum_v p^G(v)a_v+c/m_G\}$,
and let $G^\circ$ be a minimizing pool.
Taking the full pool gives
\begin{align*}
U_h(s)
&\le c+\sum_v w(v)\bigl(v_h(s)+\theta_h(v)\bigr) \le 1+c\left(1+\sum_{j=h+1}^H|\mathcal S_j|\right) \le 1+cS.
\end{align*}
Since $a_v\ge0$, we also have
$c/m_{G^\circ}\le U_h(s)$.
Hence
$\lambda/m_{G^\circ}
\le \lambda U_h(s)/c
\le \lambda/c+\lambda S
\le1$,
where the last inequality follows from
$c\ge2\lambda$ and $\lambda\le1/(2S)$.
Thus, $p^{G^\circ}\in\mathcal P_\lambda(w)$.

Evaluating the Bellman minimum at this feasible comparison
distribution yields
\begin{align*}
\Psi_h(s)
&\le
\mathcal F_h(s,p^{G^\circ};\Psi_{h+1}) \le
R_h(s,p^{G^\circ})
+\sum_v p^{G^\circ}(v)\bigl(v_h(s)+\theta_h(v)\bigr) \\
&\le
\frac{c}{m_{G^\circ}}
+\sum_v p^{G^\circ}(v)a_v
=U_h(s).
\end{align*}
Here, the second inequality follows from the definition of
$\theta_h$, and the third uses the assumed bound on $R_h$.

By optimality of $U_h(s)$, every nonempty pool $G$ satisfies
$\sum_{v\in G}w(v)(U_h(s)-a_v)\le c$.
Maximizing the left-hand side over subsets, with the empty subset
also allowed, gives
$\sum_v w(v)[U_h(s)-a_v]_+\le c$.
Since $\Psi_h(s)\le U_h(s)$, it follows that
$\sum_v w(v)
[\Psi_h(s)-v_h(s)-\theta_h(v)]_+\le c$.

Finally, the definition of $\theta_{h-1}$ implies
$\theta_{h-1}(v)
\le\theta_h(v)
+\sum_{s\in\mathcal S_h}
[\Psi_h(s)-v_h(s)-\theta_h(v)]_+$.
Therefore,
\begin{align*}
\sum_v w(v)\theta_{h-1}(v)
&\le
\sum_v w(v)\theta_h(v)+c|\mathcal S_h| \le
c\sum_{j=h}^H|\mathcal S_j|,
\end{align*}
which completes the induction.

\paragraph{Initial value.}
Since $v_1=0$ and $R_1(s,w)\le c$, we obtain
\begin{align*}
\Psi_1(s_1)
&\le
R_1(s_1,w)+\sum_v w(v)\theta_1(v) \le
c\left(1+\sum_{h=2}^H|\mathcal S_h|\right) \le cS.
\end{align*}
Thus, $q^\star\in\mathcal Q(z)$ satisfies
$\Phi_R(z,q^\star)=\Psi_1(s_1)\le cS$,
as desired.

\end{proof}

\begin{lemma}[Drift bound with bandit feedback]\label{lem:omd-planning}
Let $\overline{\calJ}_h(z,s,q)$ be defined as in
\pref{eq:bandit-cost-IX}, and recall that
$\Phi(z,q)=\sup_{T\in\mathcal T_{\mathrm{all}}}
\bigl\{\operatorname{Drift}(z,q,T)
+\lambda\sum_{h,s}d_h^{z,q,T}(s)
\overline{\calJ}_h(z,s,q)\bigr\}$.
For any $0<\lambda\le 1/(2S)$ and $z\in\mathcal Z$
with induced weight $w$ and policy $\pi$ as in
\pref{eq:induce}, it holds that
$\min_{q\in\mathcal Q(z)}\Phi(z,q)\le 2S(A+1)\lambda$.

\noindent Moreover, the total time required to compute $q^k$ across all $K$
episodes of \pref{alg:main} is 
\[
O\!\left(S^2KN^{7/2}\log(KS)\right).
\]
\end{lemma}

\begin{proof}
We apply \pref{lem:general-drift-cost} with the local penalty
$$R_h(s,q_h(\cdot\mid s))
=\lambda\overline{\calJ}_h(z,s,q)
=2\lambda\sum_v\frac{q_h(v\mid s)^2}{w(v)}
 \left(1+\sum_{a\in\mathcal A_h}
                 \frac{\pi_h(a\mid s,v)}{\sum_{v'}q_h(v'|s)\pi_h(a\mid s,v')}\right) .$$
This is well defined because, for fixed $z$,
$\overline{\calJ}_h(z,s,q)$ depends on $q$ only through
$q_h(\cdot\mid s)$.

For any nonempty $G\subseteq\operatorname{supp}(w)$, let
$m_G=\sum_{v\in G}w(v)$ and
$p^G(v)=w(v)\Indi\{v\in G\}/m_G$.
Write $\mu_h^G(a\mid s)=\sum_v p^G(v)\pi_h(a\mid s,v)$.
Using $p^G(v)^2/w(v)=p^G(v)/m_G$ on
$\operatorname{supp}(w)$, we obtain
\begin{align*}
R_h(s,p^G)
&=
\frac{2\lambda}{m_G}
\left(
1+\sum_{a:\mu_h^G(a\mid s)>0}
\frac{\sum_v p^G(v)\pi_h(a\mid s,v)}
     {\mu_h^G(a\mid s)}
\right) \\
&=
\frac{2\lambda}{m_G}
\left(
1+\left|\left\{a\in\mathcal A_h:
\mu_h^G(a\mid s)>0\right\}\right|
\right) \\
&\le
\frac{2(A+1)\lambda}{m_G},
\end{align*}
where actions with zero sampling probability contribute zero.
Thus, the local cost condition in \pref{lem:general-drift-cost}
holds with $c=2(A+1)\lambda$.
Since $c\ge 2\lambda$ and $0<\lambda\le 1/(2S)$,
that lemma yields
$\min_{q\in\mathcal Q(z)}\Phi(z,q)
\le cS=2S(A+1)\lambda$.

\paragraph{Computational cost.}
For the above choice of $R_h$, the local optimization in
\pref{eq:optr} is 
\begin{align}
\operatorname*{arg\,min}_{p\in\mathcal P_\lambda(w)}
\left\{
R_h(s,p)+\sum_v p(v)v_h(s) +
\sum_{a\in\mathcal A_h}
\max_{s'\in\mathcal S_{h+1}}
\sum_v p(v)\pi_h(a\mid s,v)
\bigl(\Psi_{h+1}(s')-v_{h+1}(s')\bigr)\right\}.\nonumber
\end{align}

Indeed, $R_h(s,p) = 2\lambda\sum_v\frac{p(v)^2}{w(v)}
 \left(1+\sum_{a\in\mathcal A_h}
                 \frac{\pi_h(a\mid s,v)}{\sum_{v'}p(v')\pi_h(a\mid s,v')}\right)$ is a sum of convex
quadratic and quadratic-over-linear functions, while the remaining
Bellman objective consists of a linear term and maxima of affine
functions. One can verify that $R_h(s,\cdot)$ is lower semicontinuous.
Recall that $N=|\mathcal V|$. Along the iterates of Algorithm 1, all candidate weights remain strictly positive, so $|\operatorname{supp}(w^k)|=N$. 
Introducing epigraph variables yields a second-order cone program
with $O(N+|\mathcal A|)$ variables,
$|\mathcal A|+1$ cones of dimension $O(N)$, and
$O(N+|\mathcal A||\mathcal S_{h+1}|)$ linear constraints.
Standard conic optimization methods can therefore solve each
subproblem to additive accuracy $\varepsilon$ in time $O(SN^{7/2}\log(S/\varepsilon))$ (assuming $N \gg SA$)
in the input encoding length and $\log(1/\varepsilon)$.
The backward pass requires at most $S$ such solves and does not
enumerate candidate pools.
In particular, solving each local minimization to a feasible
solution with objective gap at most $\varepsilon/H$ gives an
$\varepsilon$-optimal solution to the overall planning problem. By setting $\varepsilon = O(1/K)$, the total running time is bounded by $O\left(  S^2K N^{7/2} \log(KS) \right) $.
\end{proof}

\begin{lemma}[Restatement of \pref{lem:drift}]
    Let $\overline{\calJ}_h(z,s,q)$ be defined as in
\pref{eq:bandit-cost}, and recall that
$\Phi(z,q)=\sup_{T\in\mathcal T_{\mathrm{all}}}
\bigl\{\operatorname{Drift}(z,q,T)
+\lambda\sum_{h,s}d_h^{z,q,T}(s)
\overline{\calJ}_h(z,s,q)\bigr\}$.
For any $0<\lambda\le 1/(2S)$ and $z\in\mathcal Z$
with induced weight $w$ and policy $\pi$ as in
\pref{eq:induce}, it holds that
$\min_{q\in\mathcal Q(z)}\Phi(z,q)\le 2S\lambda$.
\end{lemma}
\begin{proof}
We apply \pref{lem:general-drift-cost} with $c=2\lambda$ and the local
penalty
$$R_h(s,q_h(\cdot\mid s))
=\lambda\overline{\calJ}_h(z,s,q)
=\lambda\sum_v \frac{q_h(v\mid s)^2}{w(v)}.$$
This is well defined because, for fixed $z$,
$\overline{\calJ}_h(z,s,q)$ depends on $q$ only through
$q_h(\cdot\mid s)$.
For every $(h,s)$ and every nonempty
$G\subseteq\operatorname{supp}(w)$, we have
\begin{align*}
R_h(s,p^G)
&=\lambda\sum_{v\in G}\frac{w(v)}{m_G^2}=\frac{\lambda}{m_G}
\le \frac{c}{m_G}.
\end{align*}
Thus, the cost condition in \pref{lem:general-drift-cost} holds.
Since $c=2\lambda$ and $0<\lambda\le 1/(2S)$, the lemma yields
$\min_{q\in\mathcal Q(z)}\Phi(z,q)\le cS=2S\lambda$.
\end{proof}

\subsection{Proof of \pref{thm:main}}
\label{app:omd-regret}

\begin{proof}[Proof of \protect\pref{thm:main}]
For the realized opponent-policy sequence, let
$\mu_h^{\mathrm{emp}}(\cdot\mid s)$ attain the minimum in
\pref{eq:empirical-value} for every $(h,s)$.
We construct $v^\dagger\in\Vset$ by rounding the normalized
empirical value backward.
Let $v_{H+1}^\dagger=0$ and $v_1^\dagger(s)=0$ for
$s\in\Sset_1$.
For $h=H,\ldots,2$ and $s\in\Sset_h$, let
$v_h^\dagger(s)$ be the smallest multiple of $1/(HK)$
no smaller than
\[
\max_{j\in\{1,\ldots,K\}}\sum_{a,b}
\mu_h^{\mathrm{emp}}(a\mid s)\nu_h^j(b\mid s)
\left[
\frac{\ell_h(s,a,b)}{H}
+P_hv_{h+1}^\dagger(s,a,b)
\right].
\]

We first verify that $v^\dagger\in\Vset$.
If $0\le v_{h+1}^\dagger(s')\le(H-h)/H$ for every
$s'\in\Sset_{h+1}$, then the maximum in the definition
of $v_h^\dagger(s)$ lies in $[0,(H-h+1)/H]$.
Since $(H-h+1)/H$ is a multiple of $1/(HK)$,
upward rounding preserves this interval.
Backward induction from $v_{H+1}^\dagger=0$ therefore
shows that $v^\dagger\in\Vset$.
We next show that, for every $h\ge2$ and
$s\in\Sset_h$,
\[
0\le v_h^\dagger(s)-\frac{V_h^{\mathrm{emp}}(s)}{H}
\le\frac{H-h+1}{HK}.
\]
The claim holds at $h=H+1$, where both values are zero.
Suppose that
$0\le v_{h+1}^\dagger(s')-V_{h+1}^{\mathrm{emp}}(s')/H
\le(H-h)/(HK)$ for every $s'\in\Sset_{h+1}$.
Since $P_h$ is a probability distribution,
\begin{align*}
0\le
P_hv_{h+1}^\dagger(s,a,b)
-\frac{P_hV_{h+1}^{\mathrm{emp}}(s,a,b)}{H}
\le\frac{H-h}{HK}.
\end{align*}
By the empirical Bellman equation, replacing
$v_{h+1}^\dagger$ with $V_{h+1}^{\mathrm{emp}}/H$
in the maximum defining $v_h^\dagger(s)$ gives
$V_h^{\mathrm{emp}}(s)/H$.
The induction hypothesis shows that replacing
$V_{h+1}^{\mathrm{emp}}/H$ with $v_{h+1}^\dagger$
increases this maximum by at most $(H-h)/(HK)$,
while upward rounding adds at most $1/(HK)$.
This proves the claim by backward induction.

We now bound the Bellman residuals under
$\mu^{\mathrm{emp}}$.
For every episode $k$, $h\ge2$, and $s\in\Sset_h$,
the definition of $Y^k$ gives
\begin{align*}
\sum_a\mu_h^{\mathrm{emp}}(a\mid s)
Y_{h,s,v^\dagger,a}^k
&=\sum_{a,b}\mu_h^{\mathrm{emp}}(a\mid s)
\nu_h^k(b\mid s)
\left[
\frac{\ell_h(s,a,b)}{H}
+P_hv_{h+1}^\dagger(s,a,b)
\right]
\le v_h^\dagger(s),
\end{align*}
where the inequality follows from the maximization
over $j\in\{1,\ldots,K\}$ in the definition of
$v_h^\dagger(s)$ and upward rounding.

For $h=1$, the approximation bound at $h=2$
and the empirical Bellman equation give
\begin{align*}
&\sum_a\mu_1^{\mathrm{emp}}(a\mid s)
Y_{1,s,v^\dagger,a}^k\\
&\le\sum_{a,b}\mu_1^{\mathrm{emp}}(a\mid s)
\nu_1^k(b\mid s)
\left[
\frac{\ell_1(s,a,b)}{H}
+\frac{P_1V_2^{\mathrm{emp}}(s,a,b)}{H}
\right]
+\frac{H-1}{HK}\\
&\le\frac{V_1^{\mathrm{emp}}(s)}{H}
+\frac{H-1}{HK}.
\end{align*}

By a union bound, the conclusions of
\pref{lem:omd-learner} and \pref{lem:omd-uniform-ix}
hold simultaneously with probability at least
$1-\delta$.
On this event, both bounds apply to
$z^\dagger=z^{v^\dagger,\mu^{\mathrm{emp}}}$,
even though $v^\dagger$ and $\mu^{\mathrm{emp}}$
depend on the realized opponent-policy sequence.
Using \pref{eq:omd-feedback} and
\pref{eq:omd-uniform-ix}, we obtain
\begin{align*}
\sum_{k=1}^K\langle z^\dagger,g^k\rangle
&=\sum_{k=1}^K\sum_{h=1}^H
\frac{q_h^k(v^\dagger\mid s_h^k)}{w^k(v^\dagger)}
\sum_a\mu_h^{\mathrm{emp}}(a\mid s_h^k)
\bigl(
\widehat Y_{h,s_h^k,v^\dagger,a}^k
-v_h^\dagger(s_h^k)
\bigr)\\
&\le\sum_{k=1}^K\sum_{h=1}^H
\frac{q_h^k(v^\dagger\mid s_h^k)}{w^k(v^\dagger)}
\left(
\sum_a\mu_h^{\mathrm{emp}}(a\mid s_h^k)
Y_{h,s_h^k,v^\dagger,a}^k
-v_h^\dagger(s_h^k)
\right)\\
&\quad+\frac{\log|\Vset|+S\log A+\log(2/\delta)}
{\lambda}\\
&\le\sum_{k=1}^K\frac{V_1^{\mathrm{emp}}(s_1^k)}{H}
+\frac{H-1}{H}
+\frac{\log|\Vset|+S\log A+\log(2/\delta)}
{\lambda}.
\end{align*}
For the last inequality, the Bellman residual
$\sum_a\mu_h^{\mathrm{emp}}(a\mid s_h^k)
Y_{h,s_h^k,v^\dagger,a}^k-v_h^\dagger(s_h^k)$
is nonpositive for every $h\ge2$, and its coefficient
$q_h^k(v^\dagger\mid s_h^k)/w^k(v^\dagger)$
is nonnegative.
For $h=1$, we have $q_1^k=w^k$ and
$v_1^\dagger=0$, so the remaining contribution is
bounded by the Bellman inequality for $h=1$
established above.

Combining this bound with \pref{eq:omd-learner-bound}
and using $L^k=V_1^{\mu^k,\nu^k}(s_1^k)/H$ gives
\begin{align*}
\frac{\ENR_K}{H}
&=\sum_{k=1}^K
\left(
L^k-\frac{V_1^{\mathrm{emp}}(s_1^k)}{H}
\right)\\
&\le\frac{H-1}{H}
+\frac{C_\delta}{\lambda}
+2(A+1)S\lambda K,
\end{align*}
where $C_\delta$ is defined in \pref{thm:main}.

If $\lambda=\sqrt{C_\delta/[2(A+1)SK]}$,
substitution gives
\[
\ENR_K\le H-1+2H\sqrt{2(A+1)SKC_\delta}.
\]
Otherwise, the choice of $\lambda$ implies
$K<2(A+1)SC_\delta$, and the bound

\[
\ENR_K\le HK\le H\sqrt{2(A+1)SKC_\delta}
\]
gives the same result.
Finally,
$\log|\Vset|\le S\log(HK+1)$ by \pref{eq:grid},
yielding 
\[
\ENR_K=\otil(H^{3/2}S\sqrt{AK}).
\]

The proof is thus complete.
\end{proof}

\bibliographystyle{alpha}
\bibliography{refs}

\end{document}